\documentclass[lettersize,journal]{IEEEtran}
\usepackage{amsmath,amsfonts}
\usepackage{algorithmic}
\usepackage{array}
\usepackage[caption=false,font=normalsize,labelfont=sf,textfont=sf]{subfig}
\usepackage{textcomp}
\usepackage{stfloats}
\usepackage{url}
\usepackage{verbatim}
\usepackage{graphicx}
\usepackage{adjustbox}
\usepackage{multirow}
\usepackage{booktabs}
\usepackage{xcolor}
\usepackage{color}
\usepackage{utfsym}
\usepackage[normalem]{ulem}
\usepackage{algorithm}
\usepackage{algorithmic}
\usepackage[pagebackref,breaklinks,colorlinks]{hyperref}
\usepackage{cite}
\usepackage{natbib}
\usepackage{pdfpages}
\usepackage[normalem]{ulem}
\def\BibTeX{{\rm B\kern-.05em{\sc i\kern-.025em b}\kern-.08em
    T\kern-.1667em\lower.7ex\hbox{E}\kern-.125emX}}
\usepackage{balance}
\begin{document}

\title{Towards Human-Level 3D Relative Pose Estimation: Generalizable, Training-Free, with Single Reference}

\author{Yuan Gao*, Yajing Luo*, Junhong Wang, Kui Jia, Gui-Song Xia
\IEEEcompsocitemizethanks{
\IEEEcompsocthanksitem Y. Gao and G.-S. Xia are with the School of Artificial Intelligence, Wuhan University, Wuhan, China. E-mails: ethan.y.gao@gmail.com, guisong.xia@whu.edu.cn
\IEEEcompsocthanksitem Y. Luo is with the School of Computer Science, Wuhan University, Wuhan, China. E-mail: yajingluo@whu.edu.cn
\IEEEcompsocthanksitem J. Wang is with MoreFun Studio, Tencent Games, Tencent, Shenzhen, China. E-mail: junhongwang@tencent.com
\IEEEcompsocthanksitem K. Jia is with the School of Data Science, The Chinese University of Hong Kong, Shenzhen, China. E-mail: kuijia@cuhk.edu.cn
\IEEEcompsocthanksitem This work was supported by the National Natural Science Foundation of China (62306214, 62325111), and was also partially funded by the 2024 Shenzhen Science and Technology Major Project (202402002).
\IEEEcompsocthanksitem Corresponding authors: Yuan Gao, Gui-Song Xia.}
\thanks{* indicates equal contributions.}}

\markboth{IEEE TRANSACTIONS ON PATTERN ANALYSIS AND MACHINE INTELLIGENCE,~Vol.~0, No.~0, 2025}%
{How to Use the IEEEtran \LaTeX \ Templates}

\maketitle

\begin{abstract}
Humans can easily deduce the relative pose of a previously unseen object, without labeling or training, given only a single query-reference image pair. This is arguably achieved by incorporating i) 3D/2.5D shape perception from a single image, ii) render-and-compare simulation, and iii) rich semantic cue awareness to furnish (coarse) reference-query correspondence. 
Motivated by this, we propose a novel 3D generalizable relative pose estimation method by elaborating 3D/2.5D shape perception with a 2.5D shape from an RGB-D reference, fulfilling the render-and-compare paradigm with an off-the-shelf differentiable renderer, and leveraging the semantic cues from a pretrained model like DINOv2. Specifically, our differentiable renderer takes the 2.5D rotatable mesh textured by the RGB and the semantic maps (obtained by DINOv2 from the RGB input), then renders new RGB and semantic maps (with back-surface culling) under a novel rotated view. The refinement loss comes from comparing the rendered RGB and semantic maps with the query ones, back-propagating the gradients through the differentiable renderer to refine the 3D relative pose. As a result, \emph{our method can be readily applied to unseen objects, given only a single RGB-D reference, without labeling or training}. Extensive experiments on LineMOD, LM-O, and YCB-V show that our training-free method significantly outperforms the state-of-the-art supervised methods, especially under the rigorous \texttt{Acc@5/10/15}$^\circ$ metrics and the challenging cross-dataset settings. The codes are available at~\url{https://github.com/ethanygao/training-free_generalizable_relative_pose}.
  
\end{abstract}


\section{Introduction}
\IEEEPARstart{R}{ecent} years have witnessed great progress in 3D object pose estimation \cite{templates-pose, nguyen2023nope, wang2024object, 3DAHV, zhao2024dvmnet, wang2021nemo, kaushik2024source, gao2019estimation, gao2016exploiting, gao2016symmetric}, which estimates the 3D rotation of an object depicted in a query RGB image. As a key to facilitating interaction with real-world objects, 3D object pose estimation attracts increasing attention from various areas including computer vision, virtual/augmented reality, robotics, and human-computer interaction \cite{perez2016robot, robots, AR}. To date, the community shows great interest in generalizable 3D object pose estimation \cite{templates-pose, onepose, onepose++, relpose, 3DAHV, zhao2024dvmnet} owing to its wide applicability, which focuses on the generalization to previously unseen objects, preferably in a zero-shot manner\footnote{We discuss the relatively easier instance- or category-level object pose estimation in the Related Work Sect. \ref{instance-level} and \ref{category-level}, respectively.}.

\begin{figure}[t]
\centerline{\includegraphics[width=\linewidth,height=0.45\linewidth]{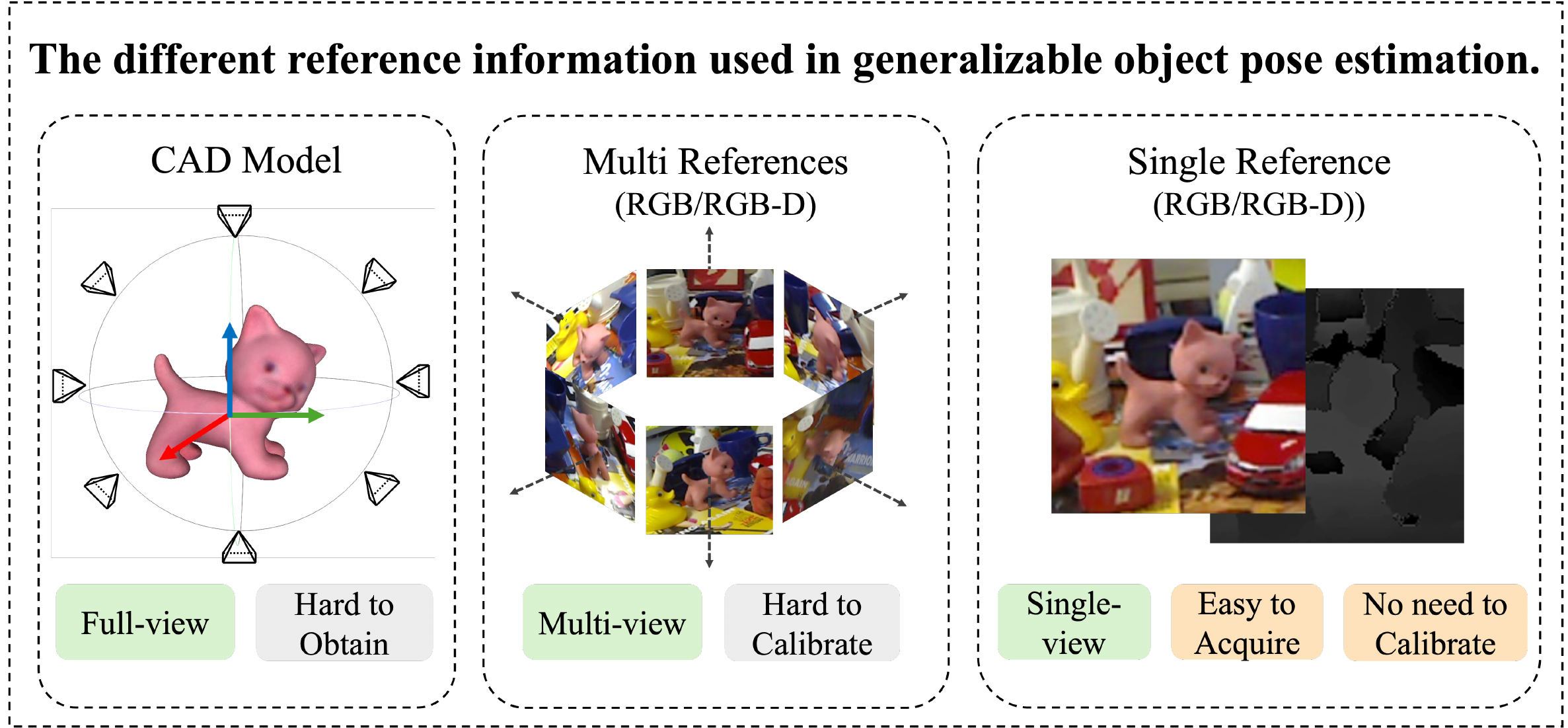}}
  \vspace{-2mm}
  \caption{\textbf{Generalizable object pose estimation with different references, i.e., \emph{a CAD model, multiple images}, or \emph{a single image}.} CAD models and multi-view references offer rich geometry details, however, scanning the precise CAD model and/or calibrating dense views for multiple images are laborious or even impossible for unseen objects in practice, such as augmented reality. We thus focus on estimating the relative pose w.r.t. a single-view reference following \cite{3DAHV, zhao2024dvmnet}, i.e., the relative pose between a reference-query pair, where we treat the reference pose as canonical without any calibration.}
  \label{fig:fig_1}
  \vspace{-4mm}
\end{figure}

\begin{table*}[t]
\vspace{-3mm}
\caption{The taxonomy of our method in \textbf{generalizable pose estimation}. For each column, we illustrate the applicability in descending order using the text of \textbf{Bold}, \underline{Underlined}, and Normal. We also include the Human Intelligence as a reference. The state-of-the-art methods used in our experiments are highlighted with their name.}
\label{tab:taxonomy}
\vspace{-0.5em}
\centering
\resizebox{0.8\textwidth}{!}{
\begin{tabular}{l cccccc}
\toprule 
\multirow{2}{*}{Method} & \multirow{2}{*}{Training} & \multirow{2}{*}{Label} & \multicolumn{2}{c}{Reference} & \multicolumn{2}{c}{Query} \\
\cmidrule(lr){4-5} \cmidrule(lr){6-7} 
& & & Modality & \#Instance & Modality & \#Instance \\
\midrule
Human Intelligence &\textbf{training-free} & \textbf{label-free} & \textbf{RGB} & \textbf{single} & \textbf{RGB} & \textbf{single}  \\

\midrule 
\cite{zeropose,zephyr,lin2023sam6d}              & supervised &  \underline{pose} & CAD & multiple & RGB-D & \textbf{single}\\
\cite{templates-pose, osop, megapose, ornek2023foundpose, nguyen2023gigapose} & supervised &  \underline{pose} & CAD & multiple& \textbf{RGB} & \textbf{single}\\
\cite{fs6d,latentfusion} & supervised & \underline{pose} & \underline{RGB-D} & multiple & RGB-D & \textbf{single}\\
RelPose++ \cite{relpose++}, \cite{gen6d, onepose, onepose++,relpose} & supervised & \underline{pose} & \textbf{RGB} & multiple & \textbf{RGB} & \textbf{single}\\
LoFTR \cite{loftr} & supervised & pose+depth & \textbf{RGB} & \textbf{single} & \textbf{RGB} & \textbf{single} \\
3DAHV \cite{3DAHV}, DVMNet \cite{zhao2024dvmnet} & supervised & \underline{pose} & \textbf{RGB} & \textbf{single} & \textbf{RGB} & \textbf{single} \\
ZSP \cite{zsp} & \textbf{training-free} & \textbf{label-free} & \underline{RGB-D} & \textbf{single} & RGB-D & multiple \\
Ours & \textbf{training-free} & \textbf{label-free} & \underline{RGB-D} & \textbf{single} & \textbf{RGB} & \textbf{single} \\

\bottomrule 
\end{tabular}}
\vspace{-1em}
\end{table*}

Existing generalizable 3D object pose estimation methods can be categorized according to how they exploit the reference information, i.e., using \emph{a CAD model, multiple images}, or \emph{a single image} as references, as shown in Fig. \ref{fig:fig_1}. Specifically, most existing methods leverage a 3D CAD model \cite{megapose, templates-pose, osop, zeropose, zephyr} or multiple images \cite{latentfusion, gen6d, onepose, onepose++, fs6d, relpose, relpose++} for template matching or feature extraction, while the requirement of laborious 3D scanning (for the CAD-based methods) or multiple-image pose labeling (for most multi-image methods) severely limits their applicability. 

On the other hand, recent methods propose to reframe the generalizable object pose estimation task as relative pose estimation between a query and a reference image from an unseen object, which is termed as \emph{generalizable \textbf{relative} object pose estimation} \cite{3DAHV,zhao2024dvmnet}. By treating the reference pose as canonical, estimating the relative pose between the reference-query pair successfully bypasses the laborious 3D scanning (of the CAD reference) or dense views calibration (of the multiple-image reference). However, existing methods rely on a large amount of well-labeled poses between the query-reference pairs to effectively train a neural network, thereby imposing the challenge of acquiring high-quantity training data \cite{3DAHV, zhao2024dvmnet, relpose, relpose++}. Moreover, the generalizability of some network-based methods may be impeded by the training data. Our empirical findings suggest that after pretrained on an external large-scale dataset such as Objaverse \cite{objaverse}, the current state-of-the-art methods \cite{3DAHV, zhao2024dvmnet} require in-dataset finetuning\footnote{The in-dataset finetuning denotes that the finetune set comes from the same dataset with the testing set, while not including the testing objects.} before testing on unseen objects within the dataset, which might potentially hinder their cross-dataset generalizability.

In this context, we work towards universally applicable zero-shot 3D generalizable relative pose estimation, where i) the object is agnostic/unseen from a cross-dataset, ii) only a single RGB-D image is available for reference without a 3D CAD model or multi-view images, and iii) the ground-truth (relative) pose label is not available. In other words, we aim to establish a novel 3D generalizable (in terms of both objects and datasets) relative pose estimation method given only one reference and one query image, without labeling or training. This is extremely challenging due to the mixture of \emph{incomplete shape information} and \emph{missing reference-query correspondence}, which leads to a severely degraded optimization problem.

Our method is inspired by the fact that humans can easily infer the relative pose under the aforementioned rigorous setting, even with large pose differences or severe occlusions. We hypothesize that such intelligence is accomplished through i) perceiving 3D/2.5D shapes from a single image, ii) conducting render-and-compare simulations via imagination, and iii) understanding rich semantic cues of the object. For example, given two viewpoints of an unseen animal, humans are able to infer the 3D/2.5D shape of that animal, then identify the correspondences of the animal eyes, noses, ears, etc, and finally rotate and render the 3D/2.5D model until its projection matches the other view. Note that the semantic cues have the potential to deal with the (self-) occluded missing parts, thus enhancing the comparison process, e.g., an animal tail can be simply ignored in the render-and-compare simulations if it only appears in one image and is (self-) occluded in the other.

The above analysis motivates us to break down our difficulties and fulfill those three requirements. Concretely, we achieve this by formulating a label/training-free framework through an off-the-shelf differentiable renderer following the render-and-compare paradigm. Our input shape to the differentiable renderer is an RGB- and semantic-textured 2.5D mesh of the reference (avoiding the difficult 3D hallucination of an unseen object). Based on this, we construct a pose refinement framework, where the differentiable renderer takes an initial pose to render projections, then back-propagates the gradients from the projection loss (between the rendered and the query images) to refine the initial pose.

Specifically, our method starts with an RGB-D reference and an RGB query, where their semantic maps can be obtained by leveraging an advanced pretrained model DINOv2 \cite{dinov2} with the RGB inputs\footnote{Note that our method possesses the potential of using only an RGB reference, please see the discussion in Sect. \ref{sec:Taxonomy} (Applicability) and Sect. \ref{sec:conclusion} (Limitations and Future Works) for more details. Moreover, our method works reasonably well even without the DINOv2 semantic maps on the LineMOD dataset, as illustrated in Table \ref{tab:ablation}.}. We leverage an easy-to-use differentiable renderer nvdiffrast \cite{nvdiffrast}, which takes the RGB- and semantic-textured 2.5D mesh of the reference as input, then renders new RGB and semantic maps (with back-surface culling) under a novel rotated view. The pose refinement loss comes from comparing the rendered RGB and semantic maps with the query ones, which flows the gradients through the differentiable renderer to refine the 3D relative pose. As a result, our method can be readily applied to unseen objects from an arbitrary dataset without labeling or training, and naturally generalizes to cross-dataset scenarios.

In summary, we propose a novel 3D generalizable relative pose estimation method, which takes only an RGB-D reference and an RGB query pair, without requiring the ground-truth pose labels or training. We achieve this by formulating a pose refinement framework via an off-the-shelf differentiable renderer under the render-and-compare paradigm. Our method does not involve training a network, which naturally possesses zero-shot generalizability in terms of both unseen objects and datasets. We conducted extensive experiments on LineMOD \cite{linemod}, LM-O \cite{lmo} and YCB-V \cite{posecnn} datasets. The results from our training-free method exhibit significant improvement over the state-of-the-art supervised methods, e.g., for \texttt{Acc@15}$^\circ$ metric on the LineMOD dataset \cite{linemod} and the YCB-V dataset \cite{posecnn}, our label- and training-free method outperforms the supervised state-of-the-art results by \textbf{29.98\%} and \textbf{14.28\%}, respectively. Our contributions are three-fold:
\begin{itemize}
    \item We propose a novel and simple relative pose estimation method that naturally generalizes to unseen objects, without the need for ground truth pose labels or neural network training.
    \item Our method is optimized by an off-the-shelf differentiable renderer and thus training-free. This eliminates the training-data dependency in supervised CNN/ViT-based methods, thus ensuring inherently robust generalizability.
    \item Our method employs a render-and-compare framework leveraging 2.5D meshes, thereby avoiding the challenging 3D hallucination of unseen objects. Building upon this, semantic features (such as those from DINOv2) can be integrated as texture maps (via PCA dimensionality reduction) into the render-and-compare process.
\end{itemize}

\vspace{-0.2cm}
\subsection{Taxonomy and Applicability of Our Method}
\label{sec:Taxonomy}

\noindent  \textbf{Taxonomy}. The taxonomy of our methods in generalizable pose estimation, in terms of \texttt{training}, \texttt{labeling}, as well as the \texttt{modality} and the \texttt{number of required instances} of the \emph{reference} and the \emph{query} images, is illustrated in Table \ref{tab:taxonomy}. Our method falls under the category of \emph{label/training-free} with a \emph{single RGB query} and a \emph{single RGB-D reference}.

\noindent \textbf{Applicability}. Among Table \ref{tab:taxonomy}, the proposed method shares the closest setting to the human intelligence on relative pose estimation that is able to generalize to unseen objects from an arbitrary dataset, with only an additional one-time-collection depth map for the reference image. 

Our method arguably possesses better applicability compared to the state-of-the-art methods summarized in Table \ref{tab:taxonomy}, as i) unlike supervised in-dataset state-of-the-art methods, our method does not require ground truth (GT) pose annotations for training, where obtaining a large number of GT poses is arguably more challenging than acquiring a single reference depth map. ii) Our method requires only a one-time reference depth annotation, which can be collected and fixed in advance. Moreover, depth sensors are commonly used in our primary application domain, i.e., robotics. We have testified in supplementary material Table S1 that our method can still deliver good estimations with an imprecise depth map, simulating noisy depth sensors. iii) Our training-free method naturally generalizes to unseen objects because it does not involve training a neural network, and thus is training-data independent. In contrast, supervised state-of-the-art methods typically perform less satisfactorily in cross-dataset scenarios, as they suffer from training-data dependency, leading to generalization issues given different training and evaluation datasets.

To further examine the potential of fully distilling human intelligence, we conducted ablation experiments in the supplementary material by substituting the GT depth with predictions from the state-of-the-art Depth Anything v2 \cite{yang2024depth}. We note that to preserve the object shape, our method requires the \textbf{\textit{metric}} depth, meaning the estimated depth $z$ and spatial dimensions $x, y$ should share the same unit of measurement (e.g., both in meters). However, as shown in Table S2-S4 in supplementary material, the off-the-shelf metric depth estimator occasionally fails to generalize across different datasets, likely due to an imprecisely recovered depth scale caused by varying camera parameters and/or diverse objects across different training and evaluation datasets. We note that a generalizable metric depth estimator would alleviate this issue, but training a generalizable metric depth estimator is beyond the scope of, and may introduce distractions to, our current focus. 

Finally, our method also incorporates the segmentation maps of both query and reference objects as input, which can be obtained by pretrained segmentation models such as SAM \cite{sam}, FastSAM \cite{zhao2023fast} and Grounded SAM \cite{ren2024grounded}. We chose not to delve into these segmentation techniques extensively either, for the same sake of minimizing potential distractions.

\section{Related work}
\subsection{Instance-level 6D Pose Estimation \label{instance-level}}

Current object pose estimation can be categorized into instance-level, category-level, and generalizable methods based on different problem formulations. For instance-level methods, there are roughly three categories: direct regression-based, correspondence-based, and refinement-based. 
Direct regression-based methods \cite{bb8,ssd6d,posecnn,yolo6d,efficientpose} predict the object pose directly through a neural network. Correspondence-based methods \cite{pvnet,pvn3d,cdpn,dpod,epos,gdrnet,sopose,zebrapose,cvpr2023keypoint,checkerpose} estimate the 2D-3D/3D-3D correspondence between the 2D images and 3D object models, followed by PnP solvers \cite{Epnp} to calculate 6D poses. Additionally, refinement-based methods \cite{deepim,rnnpose,cir} incorporate refinement-based steps to improve the prediction performance. However, instance-level methods are trained on instance-specific data and rely heavily on CAD models to render numerous training data. Consequently, their application is limited to the objects on which they were trained.

\vspace{-0.2cm}
\subsection{Category-level 6D Pose Estimation \label{category-level}}

In category-level methods, the test instances are not seen during training but belong to known categories. Most methods achieve this by either alignment or directly regressing. Alignment-based methods \cite{lee2021category, nocs, socs, spd, sgpa, dpdn} first propose a Normalized Object Coordinate Space (NOCS) \cite{nocs} as a canonical representation for all possible object instances within a category. A network is then trained to predict the NOCS maps and align the object point cloud with the NOCS maps using the Umeyama algorithm \cite{umeyama1991least} to determine the object pose. This method typically constructs the mean shape of specific categories as shape priors using offline categorical object models, and the networks are trained to learn deformation fields from the shape priors to enhance the prediction of NOCS maps.
In contrast, directly regressing methods \cite{cass, dualposenet,fsnet,vinet,cps++} avoid the non-differentiable Umeyama algorithm and often focus on geometry-aware feature extraction. For instance, CASS \cite{cass} contrasts and fuses shape-dependent/pose-dependent features to predict both the object's pose and size directly. Fs-net \cite{fsnet} leverages 3D Graph Convolution for latent feature extraction, and designs shape-based and residual-based networks for pose estimation. 
However, while category-level methods strive to address different instances within the same category, their capacity to predict the poses of objects from entirely new categories remains limited, highlighting the ongoing need to broaden the scope of object pose estimation to encompass unfamiliar objects.

\subsection{Generalizable 6D Pose Estimation}
\vspace{-0.08cm}
Generalizable algorithms aim to enhance the generalizability of unseen objects without the need for retraining or finetuning. Methods in this category can be classified as CAD-based \cite{templates-pose, osop, zeropose, zephyr, ornek2023foundpose, megapose, nguyen2023gigapose, caraffa2023object} or multi-view reference-based \cite{latentfusion, gen6d, onepose, onepose++, fs6d, bundlesdf}.

For CAD-based approaches, CAD models are often used as prior knowledge for direct feature matching or template generation. In particular, ZeroPose \cite{zeropose} performs point feature extraction for both CAD models and observed point clouds, utilizing a hierarchical geometric feature matching network to establish correspondences. Following ZeroPose, SAM-6D \cite{lin2023sam6d} proposed a two-stage partial-to-partial point matching model to construct dense 3D-3D correspondence effectively. Instead, Template-Pose \cite{templates-pose} utilizes a CAD model to generate a collection of templates and selects the most similar one for a given query image. Similarly, OSOP \cite{osop} renders plenty of templates and estimates the 2D-2D correspondence between the best matching template and the query image to solve the object pose. MegaPose \cite{megapose} proposed a coarse network to classify which rendered image best matches the query image and generate an initial pose. Subsequently, multi-view renderings of the initial pose are produced, and a refiner is trained to predict an updated pose. 

Multi-view reference-based methods can be further divided into feature matching-based and template matching-based approaches. 
For the former, multi-view reference-based feature matching methods mainly aim to establish 2D-3D correspondences between the RGB query image and sparse point cloud reconstructed by reference views or 3D-3D correspondences between the RGB-D query and RGB-D reference images. For instance, FS6D \cite{fs6d} designed a dense prototype matching framework by extracting and matching dense RGBD prototypes with transformers. After the correspondence is established, Umeyama \cite{umeyama1991least} algorithms are utilized for pose estimation. OnePose/OnePose++ \cite{onepose,onepose++} 
apply the Structure from Motion (SfM) method to reconstruct a sparse point cloud of the unseen object using all reference views. They then employ an attention-based network to predict the correspondence between 2D pixels and the reconstructed point clouds to estimate the object pose.
For the latter, multi-view references can be reviewed as templates for retrieval when plenty of views exist, or used to reconstruct the 3D object models for template rendering, similar to the CAD-based methods. Gen6D \cite{gen6d} selects the closest reference view for the query image, and then refines the pose through the 3D feature volume constructed from both the reference and query images. Notably, Gen6D requires more than 200 reference images for initial pose selection. On the contrary, LatentFusion \cite{latentfusion} reconstructs a latent 3D representation of an object to present an end-to-end differentiable reconstruction and rendering pipeline, and then estimates the pose through gradients update. Since a 3D object representation can be reconstructed utilizing the multi-view information, FoundationPose \cite{wen2023foundationpose} proposed a unified framework to support both CAD-based and multi-view supported setups. When no CAD model is available, they leverage multi-view references to build a neural implicit representation, which is then used for render-and-compare.

\vspace{-0.8mm}
\subsection{Generalizable Relative Pose Estimation}
\vspace{-1mm}
Recent methods \cite{3DAHV, zhao2024dvmnet, relpose, relpose++} highlight the importance of formulating object pose estimation as a relative pose estimation problem. Specifically, \cite{3DAHV,zhao2024dvmnet} address situations where only a single-view reference image is available. \cite{3DAHV} evidence that some state-of-the-art feature matching approaches \cite{sarlin2020superglue,loftr,zsp} fail to generate reliable correspondence between the reference-query pair, while energy-based methods \cite{relpose,relpose++} struggles to capture 3D information. Instead, 3DAHV \cite{3DAHV} introduces a framework of hypothesis and verification for generating and evaluating multiple pose hypotheses. Based on that, DVMNet \cite{zhao2024dvmnet} directly lifts the 2D image features to 3D voxel information in a hypothesis-free way, computing the relative pose estimation in an end-to-end fashion by aligning the 3D voxels.

\begin{figure*}[t]
  \centerline{\includegraphics[width=6.0in,height=2.6in]{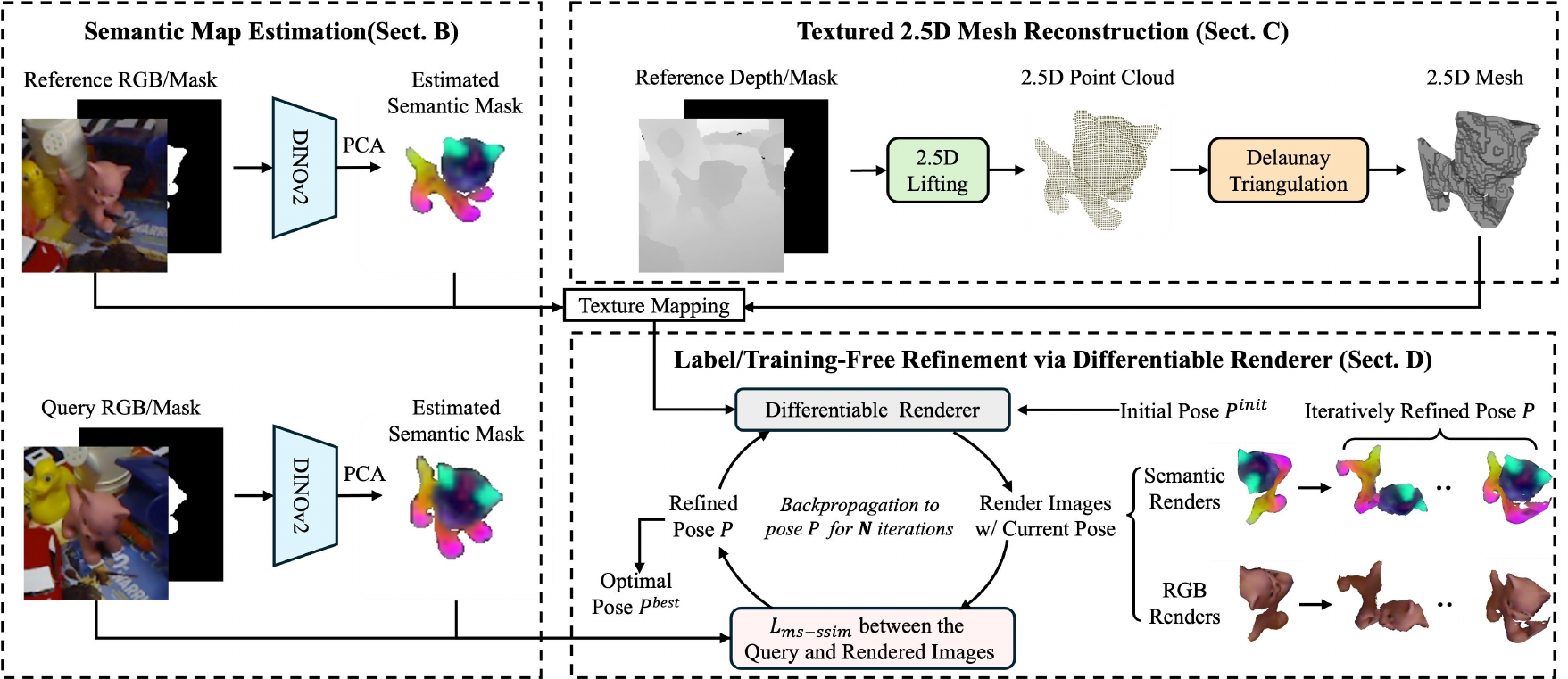}}
  \vspace{-2mm}
  \caption{\textbf{The overview of the proposed method.} Given an RGB-D reference and an RGB query, we extract the semantic maps from a pretrained DINOv2 model \cite{dinov2} for both reference and query. Then, the reference 2.5D front-surface mesh is reconstructed by the depth input without hallucination, which is subsequently texture-mapped by its RGB and semantic images. By leveraging a differentiable renderer \cite{nvdiffrast}, we generate the rendered RGB and semantic maps using the textured 2.5D reference mesh under a novel view/pose. Finally, the rendered RGB and semantic maps are compared to their query counterparts, producing losses and back-propagating the gradients through the differentiable renderer to refine the relative pose.}
  \label{fig:fig1}
  \vspace{-4mm}
\end{figure*}

\section{Method}

Following the render-and-compare paradigm, current generalizable pose estimation methods often rely on rotatable 3D CAD models or well-calibrated multi-view images, imposing challenges to acquire the 3D CAD models or expensive pose calibration, especially for previously unseen objects. We instead focus on the generalizable \textbf{relative} pose estimation formulated in the pioneering works \cite{3DAHV,zhao2024dvmnet}, which aims to estimate the relative pose between a reference-query pair, using only a single reference with an arbitrary pose as canonical (without calibration). Our method differs from \cite{3DAHV,zhao2024dvmnet} in not requiring labeled relative pose to train an estimation network.

\vspace{-1.5mm}
\subsection{Overview}

Taking an RGB query and an RGB-D reference as input, our method establishes a refinement optimization under the render-and-compare framework, by leveraging a 2.5D (i.e., RGB-D) shape of the reference, a pair of semantic maps for both the query and the reference acquired by a pretrained DINOv2 model \cite{dinov2} along with the corresponding RGB maps, and a differentiable renderer to backpropagate the gradients. Note that the 2.5D shape is exploited due to the inherent difficulty of accurately hallucinating the 3D shape of unseen objects when relying solely on a single RGB-D image. This challenge further complicates the task of relative pose estimation, as the hallucinated 3D shape must align precisely with the query to achieve a successful estimation.

Formally, by using \emph{subscript} to denote query or reference, our method starts with an RGB pair $I_r$ and $I_q$ for both reference and query, as well as a depth map $D_r$ for the reference. We proposed to estimate the relative pose between $I_r$ and $I_q$, assisted by $D_r$. To this end, we first infer the semantic maps $S_r$ and $S_q$ from $I_r$ and $I_q$ exploiting a pretrained DINOv2 model \cite{dinov2}. Then, we construct a 2.5D mesh model $M_r$ for the reference object based on $D_r$, to formulate an RGB and semantic maps textured 2.5D mesh $\mathcal{M}_r = \{M_r, I_r, S_r\}$. Subsequently, the textured 2.5D reference mesh $\mathcal{M}_r$ is rotated with an (arbitrary) initial pose $P$ by a differentiable renderer \cite{nvdiffrast} to generate novel $I_r(P)$ and $S_r(P)$. Finally, the generated $I_r(P)$ and $S_r(P)$ are compared with the query $I_q$ and $S_q$, producing a refinement loss and consequently back-propagate gradients to $P$ through the differentiable renderer. Our method operates the render-and-compare procedure in a self-supervised and network-free manner, without labeling or training.

The overview of the proposed method is illustrated in Fig. \ref{fig:fig1}. We detail the comprising elements of our method in the following sections, i.e., \textbf{semantic map estimation} in Sect. \ref{sec:sec3.1}, \textbf{textured 2.5D mesh reconstruction} in Sect. \ref{sec:sec3.2}, and \textbf{label/training-free refinement via differentiable renderer} in Sect. \ref{sec:sec3.3}.

\vspace{-0.15cm}
\subsection{Semantic Map Estimation}
\label{sec:sec3.1}

In order to estimate the relative pose, human intelligence may unconsciously infer the semantics of the reference-query pair. Subsequently, coarse correspondence can be established with those semantics, resulting in three-fold benefits: i) it helps to filter out the large non-overlapped part under a substantial pose difference, ii) alleviates the influence of occlusions, and iii) eases the degraded optimization of the relative pose estimation.

\begin{figure*}[t]
\centerline{\includegraphics[width=5in,height=1.5in]{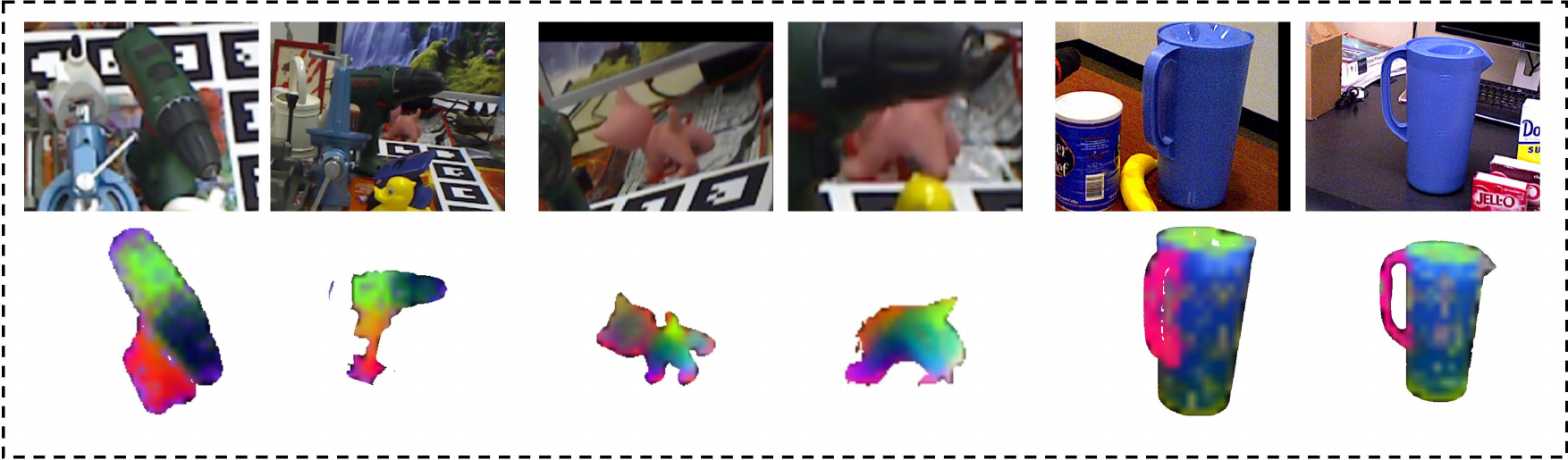}}
\caption{\textbf{Illustration of the semantic maps estimated by DINOv2 \cite{dinov2}}, demonstrating promising zero-shot performance even for texture-less objects.}
\vspace{-2mm}
\label{fig:fig2}
\end{figure*}

Benefit from the rapid development of large pretrained models, an elegant off-the-shelf semantic feature extractor is available as DINO/DINOv2 \cite{dino, dinov2}, which shows great zero-shot generalizability to diverse (even texture-less) objects (see Fig. \ref{fig:fig2} for some examples). We thus incorporate the off-the-shelf DINOv2 model \cite{dinov2} to acquire the rich semantics of the input unseen objects.

Specifically, we utilize DINOv2 \cite{dinov2} as the semantic feature extractor $\mathbf{\Phi(x)}$, which takes an RGB image $I$ to produce a set of semantic features $F \in \mathbb{R}^{w \times h \times d}$. In order to texture $F$ to the 2.5D model and facilitate the novel pose rendering, we use the principal component analysis (PCA) to reduce the dimension of $F$ from $d$ to 3, obtaining a semantic map $S$:
\begin{equation}
S = \texttt{PCA}(\mathbf{\Phi}(I)),\quad \texttt{PCA}:\mathbb{R}^{w \times h \times d} \rightarrow \mathbb{R}^{w \times h \times 3}. \label{eq:equation1}
\end{equation}
By feeding Eq. \eqref{eq:equation1} with $I_q$ and $I_r$, we can obtain the semantic maps for the query and the reference, $S_q$ and $S_r$, respectively. To ensure the semantic consistency between reference and query images, we first calculate and fix the PCA transformation using the reference image, then apply this transformation to all the query images.

\subsection{Textured 2.5D Mesh Reconstruction}
\label{sec:sec3.2}
In this section, we reconstruct a rotatable 2.5D model of the reference given its depth map $D_r$, which is subsequently used to generate novel renderings through the differentiable renderer. Note that our design avoids the challenging 3D hallucination of an unseen object from the depth map, as the hallucinated 3D shape must consistently align with the query for relative pose estimation.

Specifically, given the depth map $D_{r}$ of the reference, we lift the coordinates of the image plane into the 3D space and obtain the front surface 2.5D point clouds $X_r \in \mathbb{R}^{N \times 3}$. We then reconstruct the corresponding 2.5D mesh $M_r$ from $X_r$, to facilitate the rasterization in the renderer. Since the $xy$ coordinates of $X_r$ are sampled regularly from the 2D grids, reconstructing $M_r$ from $X_r$ can be easily achieved by the Delaunay triangulations \cite{lee1980two}. Finally, we texture $M_r$ with both color and semantic maps, obtaining $\mathcal{M}_r = \texttt{TextMap}(M_r, I_r, S_r)$ for rendering under novel poses. 

Note that as discussed in Sect. \ref{sec:Taxonomy} (Applicability), our method possesses the potential of using only an RGB reference and estimating an imprecise depth map exploiting an off-the-shelf generalizable depth estimator. Good estimation is validated in Sect. S2 of the supplementary material given an imprecise and noisy depth. We leave training a generalizable depth estimator as our future work to avoid possible distractions in this paper.

\subsection{Label/Training-Free Refinement via Differentiable Renderer}
\label{sec:sec3.3}

Our last module of label/training-free refinement is constructed by a differentiable renderer, which takes the textured 2.5D reference mesh $\mathcal{M}_r$ and a pose $P$ as input, then renders a novel RGB image and a novel semantic map under the view $P$. By implementing the pose $P$ as a random variable, the render-and-compare/reprojection loss can be back-propagated directly to $P$, ensuring the label/training-free and zero-shot unseen generalization merits of our proposed method.

Formally, by assuming a perspective camera, we leverage a recent differentiable renderer \texttt{nvdiffrast} \cite{nvdiffrast}, denoted as $\mathcal{R}$, to generate novel RGB and semantic maps, $I_r(P)$ and $S_r(P)$, from the textured 2.5D reference mesh $\mathcal{M}_r$, an arbitrary pose $P$, and the camera intrinsic $K$\footnote{The camera intrinsic $K$ can be obtained from the image EXIF information.}:
\begin{equation}
    I_{r}(P), S_r(P) = \mathcal{R}(P, \mathcal{M_\text{r}}, K)
\end{equation}

\noindent \textbf{Back Surface Culling.} As the reconstructed mesh is only 2.5D representing the front surface, it is crucial to conduct the back-surface culling during the rendering to filter out the incorrect back-facing polygons. Specifically, for every triangle of the mesh, we first calculate the dot product of their surface normal and the camera-to-triangle (usually set to $[0,0,1]$) and then discard all triangles whose dot product is greater or equal to 0 \cite{backface_culling}. Please also see the ablation with and without the back-surface culling in Table \ref{tab:ablation}.

Finally, the pose $P$ can be optimized to align the rendered $I_r(P)$ and $S_r(P)$ with the query $I_q$ and $S_q$, with the re-projection loss calculated by:
\begin{equation}
  L(P) = L_1\left\{I_{r}(P); I_{q}\right\} + L_2\left\{S_r(P); S_{q}\right\},\label{eq:equation2}
\end{equation}
where $L(P)$ is the final loss to optimize the pose $P$, and we implement both losses by the multi-scale structural similarity (MS-SSIM) \cite{msssim} as the following:
\begin{align}
  & L_1 = 1 - \text{ms-ssim}\left\{I_{r}(P); I_q\right\},  \\
  & L_2 = 1 - \text{ms-ssim}\left\{S_{r}(P); S_q\right\},\label{eq:equation5}
\end{align}
Equation \eqref{eq:equation2} enables us to optimize $P$ simply by gradient descent.

\noindent \textbf{Initialization.} As revealed in the majority of prior arts \cite{deepim, megapose, templates-pose, latentfusion}, a good initialization significantly boosts the performance of the render-and-compare framework. 

To this end, we implement our initialization by evenly sampling candidate poses on a sphere and chasing the best one. Specifically, we first sample $m$ viewpoints (azimuth and elevation angles) uniformly using a Fibonacci lattice \cite{gonzalez2010measurement}, then uniformly sample $n$ in-plane rotation angles for each viewpoint, producing $t=m*n$ poses as the initializing candidates. By rendering both RGB and semantic maps of those candidate poses, we are able to calculate the re-projection loss by Eq. \eqref{eq:equation2} (without back-propagation in this phase) and choose the pose with the minimal loss as our initialization $P^{\text{init}}$. 

Given the initialized pose $P^{\text{init}}$, we perform $N$ iterations with gradient back-propagation to carry out the label/training-free refinement via the differentiable renderer. Our algorithm is detailed in Algorithm \ref{alg:dr}.

\begin{algorithm}[!t]
\renewcommand{\algorithmicrequire}{\textbf{Input:}}
\renewcommand{\algorithmicensure}{\textbf{Output:}}
\newcommand{\rcomment}[1]{\hfill\(\triangleright\) #1}
\caption{Generalizable Label/Training-Free Refinement} 
\label{alg:dr} 
\begin{algorithmic}[1]
    \REQUIRE Reference RGB and depth $I_r, D_r$; query RGB $I_q$; differentiable renderer $\mathcal{R}$; pretrained DINOv2 model $\mathbf{\Phi}$, iteration quota $N$, learning rate $\alpha$, camera intrinsic $K$.
    
    \STATE  $S_q \leftarrow \texttt{PCA}(\mathbf{\Phi}(I_q))$, $S_r \leftarrow \texttt{PCA}(\mathbf{\Phi}(I_r))$ 
    
    \STATE $M_r \leftarrow \texttt{DelaunayTriangulations}(I_r, D_r)$ 
    \STATE $\mathcal{M}_r \leftarrow \texttt{TextMap}(M_r, I_r, S_r)$ 
    
    \vspace{1.5mm}
    $\triangleright$ \textcolor{teal}{Sampling Poses for Initialization}
    \STATE    $\left\{P^{\text{1}},P^{\text{2}},...,P^{n}\right\} \leftarrow \texttt{Uniformly\_sampling()}$
    \STATE   $\mathbf{P} = \left\{P^{\text{1}},P^{\text{2}},...,P^{n}\right\}$  
    \STATE  $I_r(\mathbf{P}), S_r(\mathbf{P}) \leftarrow \mathcal{R}(\mathbf{P},\mathcal{M_\text{r}},K)$ 
    
    \STATE \scalebox{0.9}{$P^{\text{init}}=\mathop{\arg\min}_{P^{i} \in \mathbf{P}}{L_1\left\{I_{r}(P^{i}); I_{q}\right\}+L_2\left\{S_r(P^{i}); S_{q}\right\}}$} 
    
    \vspace{1.5mm}
    
    $\triangleright$ \textcolor{teal}{Label/Training-Free Refinement via Diff. Renderer} \\

    \STATE $P \leftarrow P^{\text{init}}$
    \FOR{$i < N$} 
    \STATE $I_r(P), S_r(P) \leftarrow \mathcal{R}(P,\mathcal{M_\text{r}},K)$ 
    \STATE $L(P) = L_1\left\{I_{r}(P); I_{q}\right\} + L_2\left\{S_r(P); S_{q}\right\}$
    \STATE $P \leftarrow \texttt{GradientDescent}(L(P), \alpha)$ 
    \ENDFOR
    \ENSURE $P$
\end{algorithmic} 
\end{algorithm}

\section{Experiments}

In this section, we extensively validate our method on benchmark datasets including the LineMOD \cite{linemod}, YCB-V \cite{posecnn}, and LineMOD-Occlusion (LM-O) \cite{lmo} datasets. We detail the experimental setup in the following.

\vspace{-0.3cm}
\subsection{Experimental Setups} \label{sec:exp_setup}

\noindent \textbf{State-of-the-art Methods for Comparison.} As shown in Table \ref{tab:taxonomy}, there does not exist a method applying the challenging setting of label/training-free and a single reference-query pair like ours. Therefore we choose the state-of-the-art methods that share the closest experimental setups, which are \textbf{ZSP} \cite{zsp}, \textbf{LoFTR} \cite{loftr}, \textbf{RelPose++} \cite{relpose++}, \textbf{3DAHV} \cite{3DAHV}, and \textbf{DVMNet} \cite{zhao2024dvmnet}. Specifically, for \textbf{ZSP}, though it was originally proposed to process multiple queries, it is able to accept one RGB-D query as input. We report its performance based on the single RGB-D query and single RGB-D reference pair. For \textbf{LoFTR}, we use its pretrained weights released by the authors \cite{loftr}. The weights of \textbf{DVMNet}, \textbf{3DAHV}, and \textbf{RelPose++} are retrained on-demand to achieve their best performance (for the details, see the following \emph{Benchmark Experiments}, and the table captions of Table \ref{tab:lm}, Table \ref{tab:ycbv} and Table \ref{tab:lmo}).

\begin{figure*}[!htbp]
\vspace{-0.4cm}
\centerline{\includegraphics[width=6in,height=1.8in]{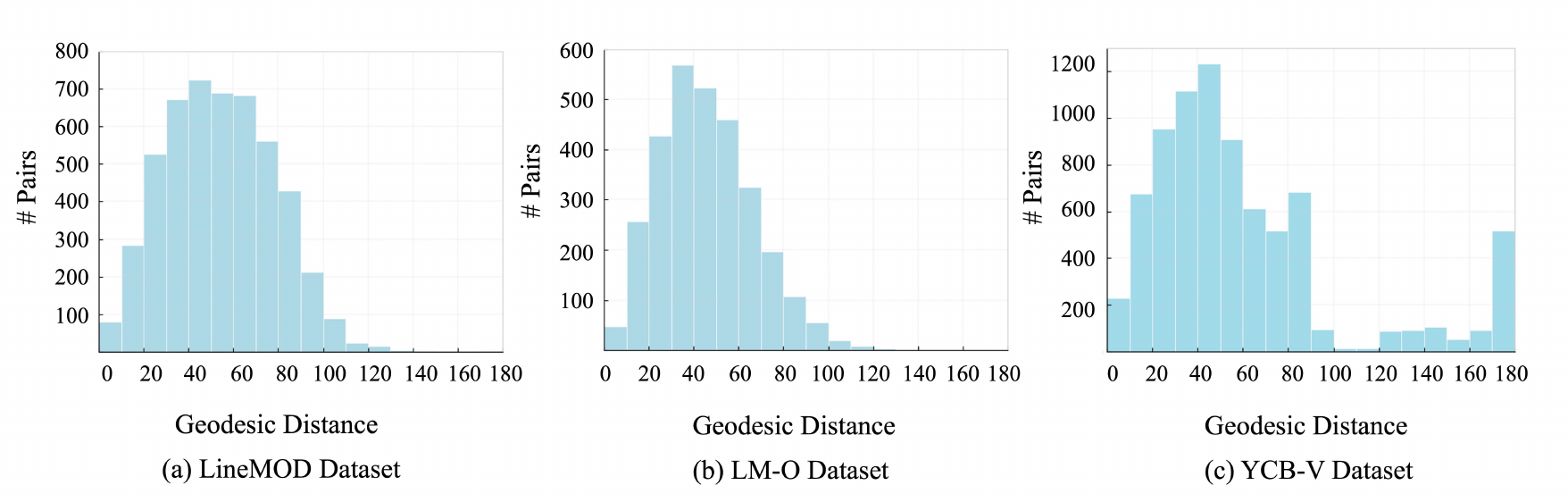}}
  \vspace{-2mm}
  \caption{\textbf{Histograms of the geodesic distance between the sampled reference-query pairs.} The in-plane rotation is included in calculating the histograms.}
  \label{fig:fighis}
  \vspace{-4mm}
\end{figure*}

\noindent \textbf{Datasets.} The experiments are carried out on three benchmark object pose estimation datasets, i.e., LineMOD dataset \cite{linemod} comprises 13 real objects, each depicting a single low-textured object on varying lighting conditions with approximately 1,200 images. LineMOD-Occlusion (LM-O) \cite{lmo} consists of 1,214 images of the 8 occluded objects, extracted from the LineMOD dataset, the average visible fraction of objects in LM-O is 79.45\%. YCB-V \cite{posecnn} encompasses over 110,000 real images featuring 21 objects characterized by severe occlusion and clutter, it exhibits an average visible object fraction of 87.15\%.

\noindent \textbf{Evaluation Metric.} Following \cite{zhao2024dvmnet,3DAHV}, we report \texttt{mean angular error} across sampled reference-query pairs. We also evaluate on important metrics of \texttt{Acc@5/10/15/30$^{\circ}$}, i.e., the percentage of the predictions that are within 5/10/15/30$^{\circ}$, which can be more rigorous (e.g., \texttt{Acc@5$^{\circ}$}) and better characterize the performance. The degree of the pose difference between the ground truth $R_{gt}$ and the predictions $\hat{R}$ is calculated by the geodesic distance $D$:
\begin{equation}
D = \arccos\left((\text{tr}(\Delta R^T_{gt} \Delta \hat{R}) - 1)/2\right) / \pi
\label{eq:equation3}
\end{equation}

\noindent \textbf{Benchmark Experiments.} The in-dataset networks of the state-of-the-art DVMNet, 3DAHV, and RelPose++ methods need to be trained on the leave-out subset which comes from the same dataset as the testing subset but does not include the testing objects. For a fair comparison, on the LineMOD dataset, we follow the experiments in DVMNet \cite{zhao2024dvmnet} and 3DAHV \cite{3DAHV} to evaluate 5 objects (i.e., benchvise, camera, cat, driller, duck). For the YCB-V experiments, we design a similar training protocol to enable the comparison with DVMNet, 3DAHV, and RelPose++, where we randomly sample 8 objects (i.e., tuna\_fish\_can, pudding\_box, banana, pitcher\_base, mug, power\_drill, large\_clamp, foam\_brick) for evaluation, leaving the remaining 13 objects to train these three methods. Following DVMNet \cite{zhao2024dvmnet}, we evaluate 3 unseen objects on the LM-O dataset (i.e., cat, driller, and duck). Since the challenging LM-O dataset is typically used solely for evaluation, we directly use the same weights for DVMNet and 3DAHV that were trained in the LineMOD experiments.

Since the results on the rigorous metrics of \texttt{Acc@5/10$^{\circ}$} are not reported in the 3DAHV \cite{3DAHV} and DVMNet \cite{zhao2024dvmnet} paper, we thus retrain them using their official codes for the \texttt{Acc@5/10$^{\circ}$} evaluation.

Moreover, as a label/training-free method, the performance of our method can be assessed on all the objects of LineMOD, YCB-V, and LM-O datasets, without the need to leave out any training data or leverage any external dataset. We report the performance of our method on the complete LineMOD, YCB-V, and LM-O datasets in Tables S5, S6, and S7 of the supplementary material.

\noindent \textbf{In-dataset and Cross-dataset Evaluation.} Beyond the unseen objects generalization, we also test the dataset-level generalization for the network-based methods including the state-of-the-art DVMNet \cite{zhao2024dvmnet} and 3DAHV \cite{3DAHV}, reporting both the \textit{in-dataset} and the \textit{cross-dataset} performance. In short, \textit{in-dataset} and \textit{cross-dataset} differ in whether the network needs to be finetuned on a subset that comes from the same dataset with the testing set (though not including the testing objects). Therefore, a good \textit{cross-dataset} performance is desirable, as the network only needs to be (pre-) trained once on a large-scale external dataset without finetuning. 

Specifically, for the \textit{in-dataset} experiments, we follow the exact training protocols of DVMNet \cite{zhao2024dvmnet} and 3DAHV \cite{3DAHV}, which first pretrain on an external large-scale dataset Objaverse \cite{objaverse} then finetune on a certain dataset (e.g., LineMOD or YCB-V). For \textit{cross-dataset} experiments, we use the pretrained weights from Objaverse directly without finetuning.

Note that our method, ZSP \cite{zsp}, and LoFTR \cite{loftr} do not require a finetuning phase before evaluation, suggesting that our method, ZSP, and LoFTR naturally generalize to an arbitrary dataset\footnote{This is achieved by that i) the pose estimation phase of our method, ZSP, and LoFTR are general and do not involve learning a network, and ii) they all use generalizable feature extractors, i.e., DINOv2 or LoFTR}.

\noindent \textbf{Reference-Query Pair Generation.} We follow DVMNet \cite{zhao2024dvmnet} and 3DAHV \cite{3DAHV} to generate the reference-query pairs with sufficient overlaps for training and testing. 
Specifically, given a reference rotation $R_{r}$ and a query rotation $R_{q}$, we first convert the rotation matrices $R_{r}$ and $R_{q}$ to Euler angles $(\alpha_{r}, \beta_{r}, \gamma_{r})$ and $(\alpha_{q}, \beta_{q}, \gamma_{q})$. Since the in-plane rotation $\gamma$ does not influence the overlaps between the reference and query pair, it is set to 0 and converted back to the rotation matrix, i.e., $\tilde{R} = h(\alpha, \beta, 0)$ with $h$ being Euler-angle to rotation matrix transformation. The overlap between the query and the reference is measured by the geodesic distance (i.e., the pose difference in degree) between their in-plane-omitted rotation matrices $\tilde{R_{q}}$ and $\tilde{R_{r}}$ using Eq. \eqref{eq:equation3}. Finally, following DVMNet \cite{zhao2024dvmnet} and 3DAHV \cite{3DAHV}, we select the sampled pairs with $\tilde{D}$ less than $90^{\circ}$.

Following DVMNet \cite{zhao2024dvmnet} and 3DAHV \cite{3DAHV}, for each object, we generate 1000 pairs for testing, and 20000 pairs for training DVMNet, 3DAHV, and RelPose++. Fig. \ref{fig:fighis} illustrates the histograms depicting the statistics of the pairwise pose difference (geodesic distance between rotation matrices $R_{r}$ and $R_{q}$) on the three datasets. All the experiments are carried out on the same testing reference-query pairs.

\noindent \textbf{Implementation Details.} For semantic feature extraction, we employ the output tokens from the last layer of the DINOv2 ViT-L model \cite{dinov2}. We use nvdiffrast \cite{nvdiffrast} as our differentiable renderer. We uniformly sample $m=200$ viewpoints and $n=20$ in-plane rotations (resulting in 4000 initialization candidates), the maximal iteration number for differentiable rendering is set to $N=30$. To backpropagate the refinement losses, we use an Adam optimizer \cite{adam} of 0.01 initial learning rate and decay by a \texttt{ReduceLROnPlateau} scheduler. All the experiments are conducted on a single NVIDIA 4090 GPU.

\begin{table*}[t]
\renewcommand{\arraystretch}{1.1} 
  \caption{\textbf{Experimental results on LineMOD.} We illustrate both the experimental settings and the performance. In the \textbf{RGB-D} category, \textbf{both} means requiring RGB-D image for both query and reference. \texttt{Acc@}$t^\circ$ measures the percentage of the estimated pose within $t^\circ$ w.r.t. the ground-truth.} 
  \vspace{-0.25cm}
  \label{tab:lm}
  \centering
  \resizebox{0.7\textwidth}{!}{
  \begin{tabular}{l||ccc||c|cccc}
    \hline
    \hline
   \multirow{2}{*}{Method} & \multicolumn{3}{c||}{Settings} & Error↓ & \multicolumn{4}{c}{Acc @ $t^{\circ}$ (\%) ↑}\\
    \cline{2-9}
    & Training & Label & RGB-D & Mean Err & 30$^{\circ}$ & 15$^{\circ}$ & 10$^{\circ}$ & 5$^{\circ}$ \\
    \hline
    ZSP       &{\usym{2717}} & label-free & both & 102.33 & 8.20 & 2.22 & 0.90 & 0.18 \\
    LoFTR     &{\checkmark} & pose+depth & no &  63.88 & 23.94 & 10.80 & 6.82 & 2.42 \\
    RelPose++ &{\checkmark} & pose & no &  46.60 & 42.50 & 15.80 & --   & --  \\
    \hline
    3DAHV (cross-dataset) &{\checkmark} & pose & no & 69.24 & 21.20 & 5.52 &  2.52  & 0.44   \\
    3DAHV (in-dataset)    &{\checkmark} & pose & no &  42.77 &  59.16 &  25.92 &  11.36  & 2.16   \\
    \hline
    DVMNet (cross-dataset) &{\checkmark} & pose & no & 47.47 & 36.44 & 13.14 & 5.92  & 1.08   \\
    DVMNet (in-dataset)    &{\checkmark} & pose & no &   33.28 & 55.02  & 22.38  & 10.66  &  2.72  \\
    \hline
    Ours (init. only)      &{\usym{2717}} & label-free & reference &  \underline{{32.24}} &  \underline{70.88} &  \underline{48.28} &  \underline{29.76} &  \underline{6.66}  \\
    Ours (init. + refine)     &{\usym{2717}} & label-free & reference &  \textbf{\underline{{29.93}}} & \textbf{\underline{72.06}} & \textbf{\underline{54.90}} & \textbf{\underline{42.74}} & \textbf{\underline{24.32}}  \\
    
    \hline
    \hline
  \end{tabular}}
\vspace{-0.2cm}
\end{table*}

\begin{figure}[t]
\vspace{-0.1cm}
\centerline{\includegraphics[width=\linewidth, height=0.28\textwidth]{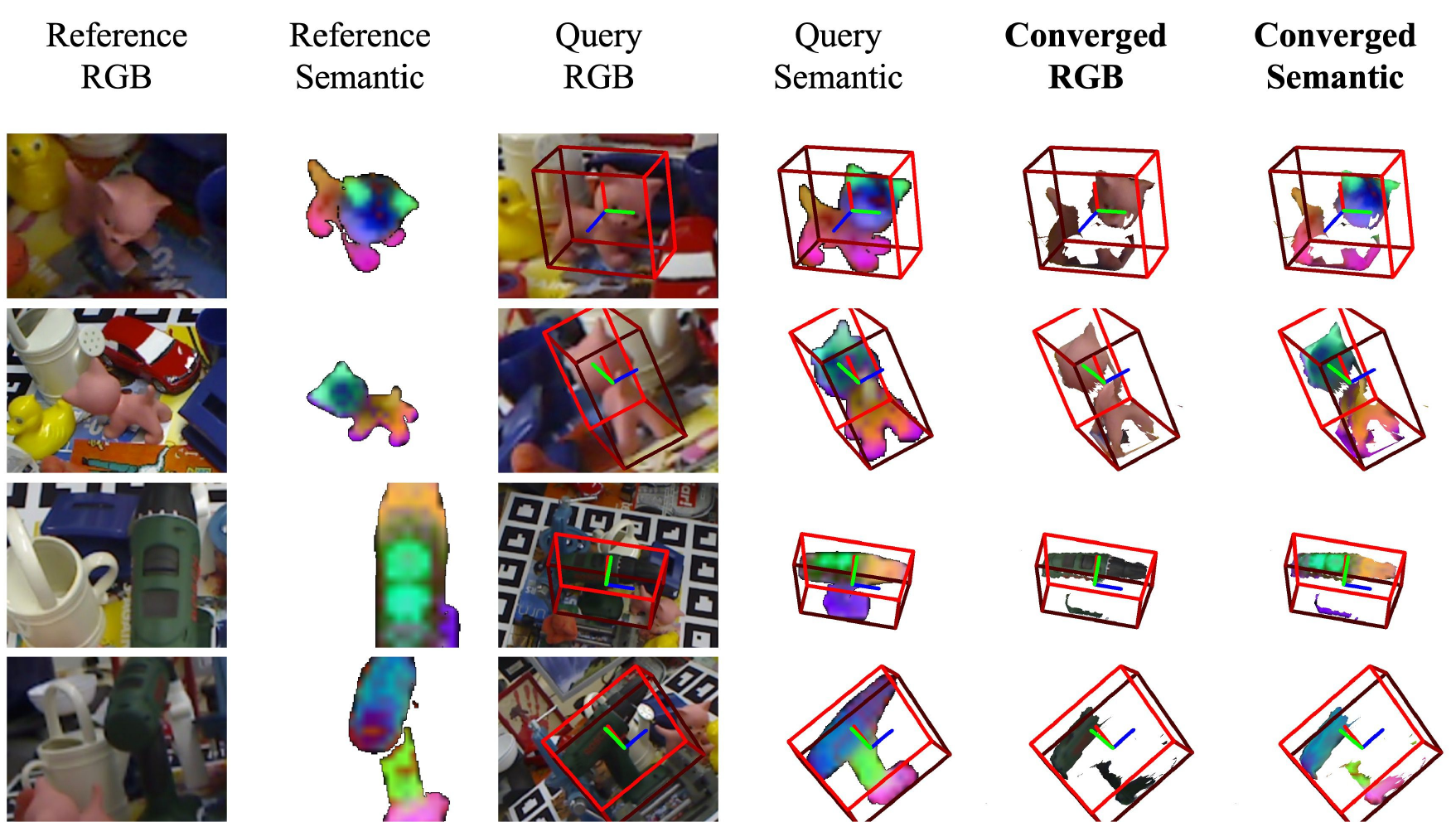}}
\caption{\textbf{Qualitative results on LineMOD.} This figure shows that our method can handle partially occluded and texture-less objects. We use a 3D bbox to denote poses.}
\label{fig:fig_vis_lm}
\vspace{-0.25cm}
\end{figure}

\subsection{Experimental Results on the LineMOD Dataset}
The results on the LineMOD dataset are illustrated in Table \ref{tab:lm}.  We paste the performances of RelPose++ from the 3DAHV paper \cite{3DAHV}. We leave the \texttt{Acc@5/10}$^\circ$ performance of RelPose++ blank as those were not reported in \cite{3DAHV} and the (pre-) training code of RelPose++ on the external large-scale Objaverse dataset is not available. Table \ref{tab:lm} shows that our \emph{label and training-free} method significantly outperforms the \emph{supervised} state-of-the-art DVMNet w.r.t. all the metrics. In addition, the experimental results show that the performance of DVMNet and 3DAHV decreases when facing cross-dataset scenarios. In contrast, without training a network, our approach inherently generalizes across diverse datasets. Especially, our method significantly outperforms DVMNet (in-dataset) for 21.6\% and 32.08\% w.r.t. the rigorous \texttt{Acc@5/10}$^\circ$. The qualitative results of our method are shown in Fig. \ref{fig:fig_vis_lm}, and comparisons with different methods are presented in Fig. S3 of the supplementary material. Our results on all the LineMOD objects are detailed in Table S5 of the supplementary material.

\subsection{Experimental Results on the YCB-V Dataset}

To compare with the state-of-the-art DVMNet \cite{zhao2024dvmnet}, 3DAHV \cite{3DAHV} and RelPose++ \cite{relpose++}, we follow the protocols discussed in Sect. \ref{sec:exp_setup} (In-dataset and Cross-dataset Evaluation) to obtain the in-dataset and cross-dataset performance of DVMNet \cite{zhao2024dvmnet} and 3DAHV \cite{3DAHV}, while RelPose++ is trained on the YCB-V dataset only. The performance on the YCB-V dataset is reported in Table \ref{tab:ycbv}, where our method exhibits a significant improvement of 11.02\% and 17.83\% w.r.t. the state-of-the-art DVMNet (in-dataset), respectively on the challenging \texttt{Acc@5/10}$^\circ$ metrics. We showcase the qualitative results of our method on the YCB-V dataset in Fig. \ref{fig:fig_ycbv_vis}, and those across different methods can be found in Fig. S5 of the supplementary material. Our results on all the YCB-V objects are shown in Table S7 of the supplementary material.

\begin{table*}[!htbp]
\renewcommand{\arraystretch}{1.1} 
  \caption{\textbf{Experimental results on YCB-V.} The performance of DVMNet, 3DAHV, and RelPose++ is obtained by training on a leave-out subset of 13 objects. Other parameters/symbols are the same as those in Table \ref{tab:lm}.} 
  \label{tab:ycbv}
  \centering
  \resizebox{0.7\textwidth}{!}{
  \begin{tabular}{l||ccc||c|cccc}
    \hline
    \hline
   \multirow{2}{*}{Method} & \multicolumn{3}{c||}{Settings} & Error↓ & \multicolumn{4}{c}{Acc @ $t^{\circ}$ (\%) ↑}\\
    \cline{2-9}
    & Training & Label & RGB-D & Mean Err & 30$^{\circ}$ & 15$^{\circ}$ & 10$^{\circ}$ & 5$^{\circ}$ \\
    \hline
    ZSP       &{\usym{2717}} & label-free  & both & 88.65 & 15.63  & 5.82 & 2.89 & 0.65    \\    
    LoFTR     &{\checkmark} & pose+depth & no & 68.65 & 29.45 & 13.56 & 7.9 & 3.19 \\
    RelPose++ &{\checkmark} & pose & no  & 57.41 & 23.60 & 7.13  & 3.28  & 0.76 \\
    \hline
    3DAHV (cross-dataset) &{\checkmark} & pose & no & 66.61 & 35.06 & 16.18 &  8.28  & 1.50   \\
    3DAHV (in-dataset)  &{\checkmark} & pose & no & 69.48 & 44.54 & 28.41 &  16.29  & 3.59   \\
    \hline
    DVMNet (cross-dataset)     &{\checkmark} & pose & no & 54.12 & 41.28 & 17.11 & 9.35  & 2.53   \\
    DVMNet (in-dataset)     &{\checkmark} & pose & no & 48.88 & 51.71 & 27.04 & 14.03 & 3.16   \\
    \hline
    Ours (init. only)      &{\usym{2717}} & label-free & reference &  \underline{48.65} & \underline{56.59} & \underline{35.68} & \underline{21.86} & \underline{5.36}  \\
    Ours (init. + refine)  &{\usym{2717}} & label-free  & reference & \textbf{\underline{47.09}} & \textbf{\underline{56.63}} & \textbf{\underline{42.69}} & \textbf{\underline{31.86}} & \textbf{\underline{14.18}} \\
                
    \hline
    \hline
  \end{tabular}}
  \vspace{-0.2cm}
\end{table*}
\vspace{-0.2cm}

\subsection{Experimental Results on the LM-O Dataset}
\begin{table*}[!htbp]
\renewcommand{\arraystretch}{1.2} 
  \caption{\textbf{Experimental results on LM-O.} LM-O is typically used solely for testing with only 8 objects under severe occlusions. The results of DVMNet and 3DAHV are tested directly using the weights trained on LineMOD. Since the model weights of RelPose++ used in Table \ref{tab:lm} were not released, we do not compare our method with RelPose++ in this experiment. Other parameters/symbols are the same as those in Table \ref{tab:lm}.}
  \label{tab:lmo}
  \centering
  \resizebox{0.7\textwidth}{!}{
  \begin{tabular}{l||ccc||c|cccc}
    \hline
    \hline
   \multirow{2}{*}{Method} & \multicolumn{3}{c||}{Settings} & Error↓ & \multicolumn{4}{c}{Acc @$t^{\circ}$ (\%) ↑}\\
    \cline{2-9}
    & Training & Label & RGB-D & Mean Err & 30$^{\circ}$ & 15$^{\circ}$ & 10$^{\circ}$ & 5$^{\circ}$ \\
    \hline
    ZSP       &{\usym{2717}}& label-free & both  & 103.70 & 7.10 & 1.67  & 0.60  & 0.07  \\
    LoFTR     &{\checkmark} & pose+depth & no & 68.15  & 20.63 &  9.00  &  4.87  &  1.87 \\
    \hline
    3DAHV (cross-dataset) &{\checkmark} & pose & no & 55.05 & 32.83 & 9.47 &  4.40  & 0.53   \\
    3DAHV (in-dataset)  &{\checkmark} & pose & no & 62.30 &  40.29 & 10.57 & 3.84  & 0.57   \\
    \hline
    DVMNet (cross-dataset) &{\checkmark} & pose & no &  \underline{51.75} & 35.52 & 12.94 &  5.30  & 1.33   \\
    DVMNet (in-dataset) &{\checkmark} & pose & no & \textbf{\underline{48.55}}  & 38.62  &  14.14   &  7.37  &  1.87  \\
    \hline
    Ours (init. only)      &{\usym{2717}} & label-free & reference &  55.94 &  \underline{53.80} &  \underline{31.72} &  \underline{17.18} &  \underline{2.80}  \\
    Ours (init. + refine)  &{\usym{2717}}& label-free & reference & 55.09  & \textbf{\underline{54.50}} & \textbf{\underline{34.97}} & \textbf{\underline{23.00}} & \textbf{\underline{6.83}} \\
    
    \hline
    \hline
  \end{tabular}
}
\vspace{-0.4cm}
\end{table*}

\begin{figure}[!t]
\vspace{-0.1cm}
\centerline{\includegraphics[width=\linewidth, height=0.28\textwidth]{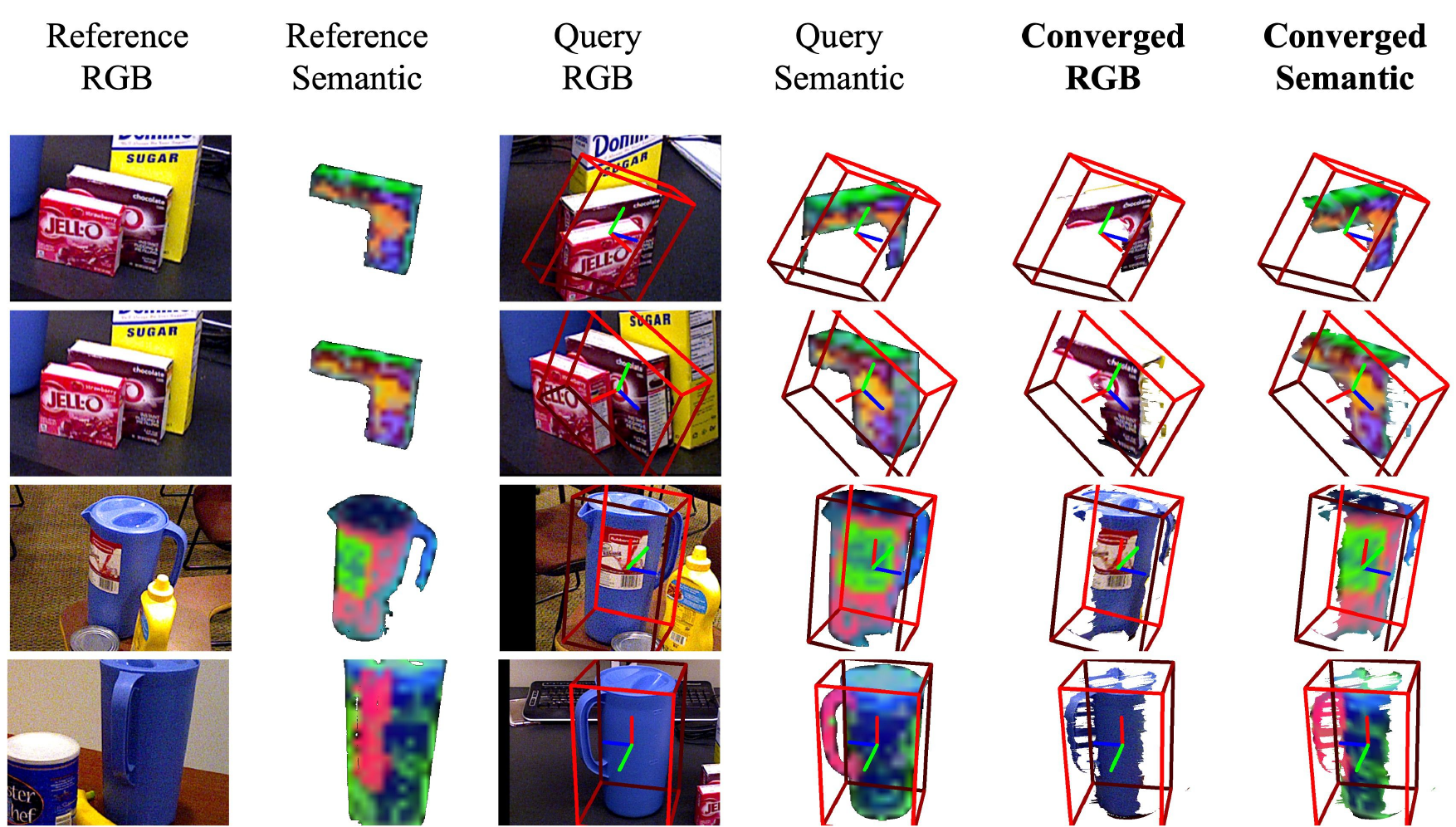}}
\caption{\textbf{Qualitative results on YCB-V.} This figure shows that our method can handle partially occluded and texture-less objects. We use a 3D bbox to denote poses.}
\label{fig:fig_ycbv_vis}
\vspace{-0.5cm}
\end{figure}

Finally, we carry out the experiments on the challenging LM-O Dataset with severe occlusions. Following DVMNet \cite{zhao2024dvmnet}, we conduct the experiments on three unseen objects of the LM-O dataset, i.e., cat, driller, and duck. We note that the LM-O dataset is typically used solely for evaluation. Therefore, the results of DVMNet and 3DAHV are evaluated utilizing the weights finetuned on LineMOD. Nevertheless, since the weights of RelPose++ for the LineMOD dataset have not been released yet and LM-O (with only 8 objects) cannot provide sufficient leave-out data to train RelPose++, we thus do not include RelPose++ for comparison. The results from Table \ref{tab:lmo} demonstrate the promising performance of our method on the severely occluded LM-O dataset. We showcase our performance on the LM-O dataset in Fig. \ref{fig:fig_lmo_vis}, and those across different methods are illustrated in Fig. S4 of the supplementary material. Our results on all the LM-O objects can be found in Table S6 of the supplementary material.

\begin{figure}[!t]
\vspace{-0.1cm}
\centerline{\includegraphics[width=\linewidth, height=0.28\textwidth]{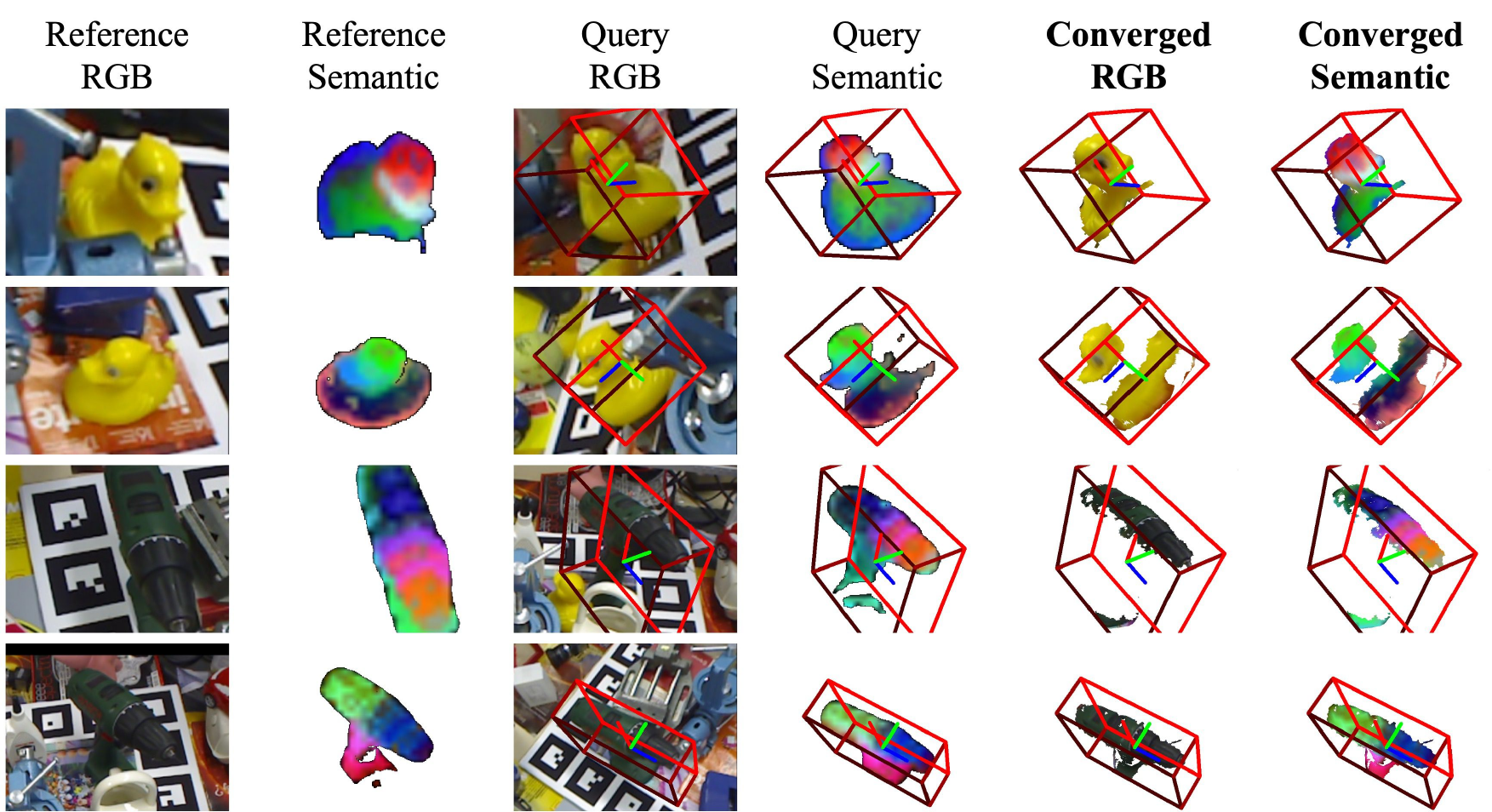}}
\caption{\textbf{Qualitative results on LM-O.} This figure shows that our method can handle severely occluded and texture-less objects. We use a 3D bbox to denote poses.}
\label{fig:fig_lmo_vis}
\vspace{-0.5cm}
\end{figure}

We observe that our results in terms of \emph{Mean Err} are inferior to the in-dataset results of the state-of-the-art DVMNet and 3DAHV (though our method exhibits better Acc@$t^{\circ}$ results). This can be attributed to the extensive occlusions presented in the LM-O dataset, which lead to numerous testing pairs lacking adequate overlap. Consequently, those testing pairs are difficult to handle by all the methods (and also challenging for humans). We show those samples as failure cases in Fig. \ref{fig:fig7} of Sect. \ref{sec:failure}, as well as investigating the angle error distribution (ranging from 0 to 180 degrees) on the LM-O dataset in Fig. S1 of the supplementary materials. The statistics reveal that at lower angle error thresholds (e.g., for $t \leq 10, 20$ in Acc@$t^{\circ}$), our approach substantially outperforms both DVMNet and 3DAHV. This indicates that for test pairs with sufficient overlaps (i.e., match-able testing pairs), our method delivers superior performance compared to the state-of-the-art DVMNet and 3DAHV.

\section{Ablation Analysis}
\begin{table}[!t]
\renewcommand{\arraystretch}{1.1} 
\vspace{-0.35cm}
  \caption{The contributions of \textbf{the proposed comprising elements} on the LineMOD dataset.}
  \vspace{-0.1cm}
  \label{tab:ablation}
  \centering
  \resizebox{\linewidth}{!}{
  \begin{tabular}{@{}l@{\hspace{2em}}c@{\hspace{1em}}c@{\hspace{1em}}c@{\hspace{1em}}c@{\hspace{1em}}c@{\hspace{1em}}c@{}}
    \hline
    \hline
     Metrics & Mean Err↓ & Acc @$30^{\circ}$↑ & Acc @$15^{\circ}$↑ & Acc @$10^{\circ}$↑ & Acc @$5^{\circ}$↑ \\
    \hline
    w/o culling  & 38.09&   67.46&  52.32   & 40.82 &   23.58 \\
    only RGB & 36.26 & 67.42 & 50.40 & 37.70 & 19.62 \\
    only semantic & 31.31   & 69.32 & 50.86 & 38.80 &   19.22 \\
    \hline
    Ours & \textbf{\underline{{29.93}}} & \textbf{\underline{72.06}} & \textbf{\underline{54.90}} & \textbf{\underline{42.74}} & \textbf{\underline{24.32}}  \\
  \hline
  \hline
  \end{tabular}
  }
  \vspace{-0.4cm}
\end{table}

\begin{table}[!t]
\renewcommand{\arraystretch}{1.1}
\caption{Ablation analysis of different \textbf{semantic features} on the LineMOD dataset.}
\vspace{-0.1cm}
\label{tab:semantic_ablation}
\centering
\resizebox{\linewidth}{!}{
\begin{tabular}{lccccc}
\hline
\hline
Metrics         & Mean Err↓ & Acc @$30^{\circ}$↑ & Acc @$15^{\circ}$↑ & Acc @$10^{\circ}$↑ & Acc @$5^{\circ}$↑ \\
\hline
RGB     & 36.26     & 67.42              & 50.40              & 37.70               & 19.62              \\
\hline
LoFTR  & 54.23 & 46.62	& 25.00 &	15.26 & 5.24 \\
RGB + LoFTR      & 39.63     & 64.30              & 45.14              & 32.42               & 14.80              \\
\hline
SD & 38.45 & 57.94 & 37.28 &26.32 & 12.12 \\
RGB + SD         & 33.78     & 65.72              & 47.56              & 36.76               & 19.90              \\
\hline
DINOv2  & 31.31     & 69.32              & 50.86              & 38.80               & 19.22              \\
RGB + DINOv2           & \textbf{\underline{29.93}} & \textbf{\underline{72.06}} & \textbf{\underline{54.90}} & \textbf{\underline{42.74}} & \textbf{\underline{24.32}} \\
\hline
\hline
\end{tabular}
}
\vspace{-0.3cm}
\end{table}

\begin{figure}[!t]
\centering
\centerline{\includegraphics[width=\linewidth]{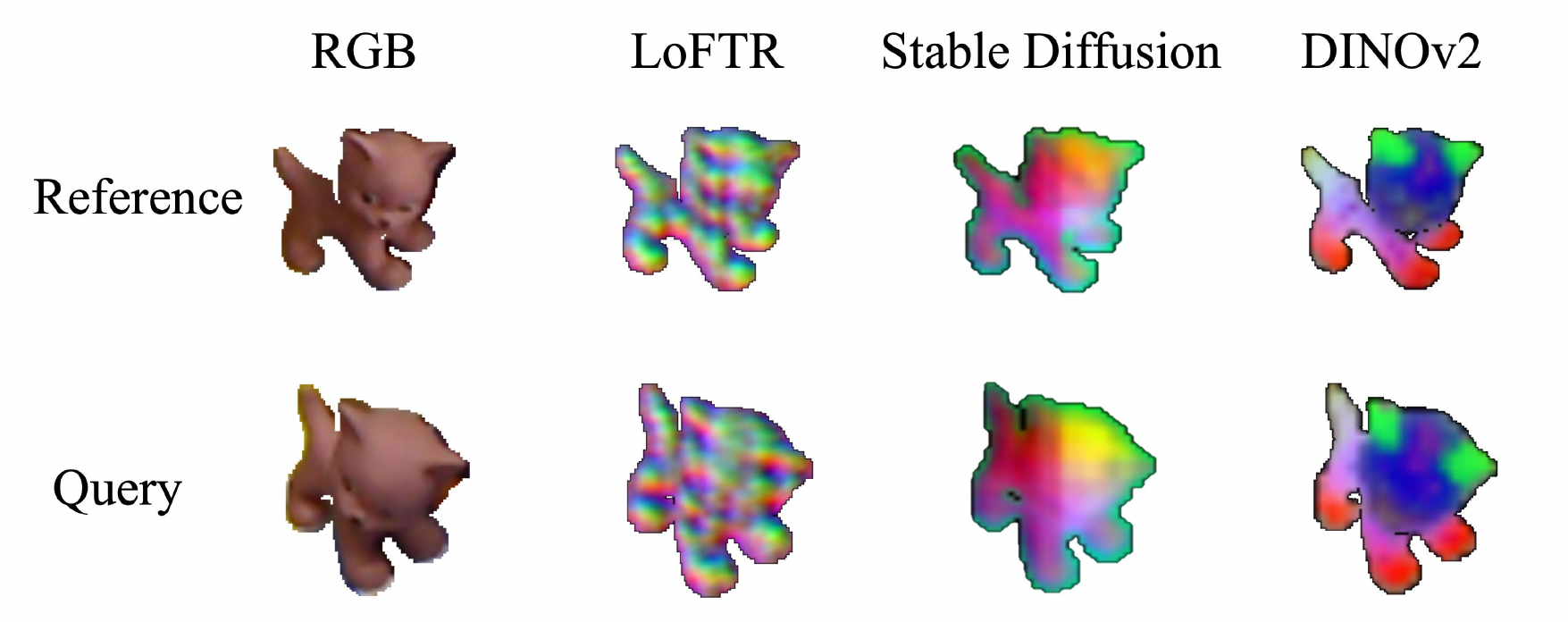}}
\vspace{-2mm}
\caption{Illustration of different semantic features as texture maps.}
\vspace{-4mm}
\label{fig:vis_pca_feature}
\end{figure}

We carefully investigate the following issues by ablation: i) the contribution of each comprising element of our method, including \emph{the back-surface culling}, and the usage of \emph{RGB} or \emph{semantic} modality in Sect. \ref{sec:component}; ii) the ablations on different semantic features in Sect. \ref{sec:semantic_features}; iii) the effects of different initialization strategies in Sect. \ref{sec:init}; iv) the effects of different refinement iterations in Sect. \ref{sec:iter}; v) the inference time statistics of our method and comparison with other baselines in Sect. \ref{sec:time}; and vi) the failure cases illustrations from the LM-O dataset in Sect. \ref{sec:failure}.

\vspace{-0.2cm}
\subsection{The Contributions of the Proposed Comprising Elements} 
\label{sec:component}

Despite the simplicity of our method, we are interested in investigating the influences for each of our comprising elements, namely \emph{the back-surface culling}, and the usage of \emph{RGB} or \emph{semantic} modality. We perform those ablations on the LineMOD, and the results are reported in Table \ref{tab:ablation}. 

As expected, removing each of our comprising elements results in a decreased performance, because all of them are exploited with clear motivations. Nonetheless, the encouraging observation is that our method is able to deliver promising results using only the \textbf{RGB} modality without the \textbf{semantic} map, which already outperforms the state-of-the-art \textbf{DVMNet (in-dataset)} \cite{zhao2024dvmnet} in Table \ref{tab:lm} across the rigorous Acc @$5^{\circ}, 10^{\circ}, 15^{\circ}$, and $30^{\circ}$. This further illustrates the good applicability of our method when the pretrained DINOv2 model is not available.

\vspace{-0.2cm}
\subsection{Ablation on Semantic features}
\label{sec:semantic_features}
To further investigate the performance incorporating alternative semantic feature representations, we tested semantic features from LoFTR \cite{loftr} and Stable Diffusion (SD) \cite{stablediffusion}. Table \ref{tab:semantic_ablation} shows that i) LoFTR and SD features are inferior to DINOv2 for this task; ii) RGB-only performance surpasses the results based solely on LoFTR or SD features; iii) complementing LoFTR, SD, or DINOv2 with RGB improves the final performance.

Table \ref{tab:semantic_ablation} reveals a significant performance gap among DINOv2, SD, and LoFTR features. To further investigate this, we visualize these features in Fig. \ref{fig:vis_pca_feature}, where the last three columns are dimensionality-reduced texture maps using PCA. Figure \ref{fig:vis_pca_feature} shows that the DINOv2 feature maps best characterize the semantic cues, well complementing the RGB appearance, thus explaining its superior performance.

\vspace{-0.2cm}
\subsection{Effects of Different Initialization Strategies}
\label{sec:init}
The pose estimation performance under the render-and-compare paradigm is largely affected by the initialization \cite{deepim, cir, rnnpose, megapose, templates-pose, latentfusion}. In the following, we investigate different initializations including: i) \emph{random initialization}, where we randomly sample candidate poses and choose the best one; and ii) \emph{uniform initialization}, where the candidate poses are uniformly sampled from a Fibonacci lattice with in-plane rotations \cite{gonzalez2010measurement}, as detailed in Sect. \ref{sec:sec3.3} (Initialization). Table \ref{tab:ablation_initial} shows the performance of different initialization strategies using the LineMOD dataset, which demonstrates that the \emph{uniform initialization} outperforms the \emph{random initialization}. 

\begin{figure}[!t]
\centerline{\includegraphics[width=0.72\linewidth]{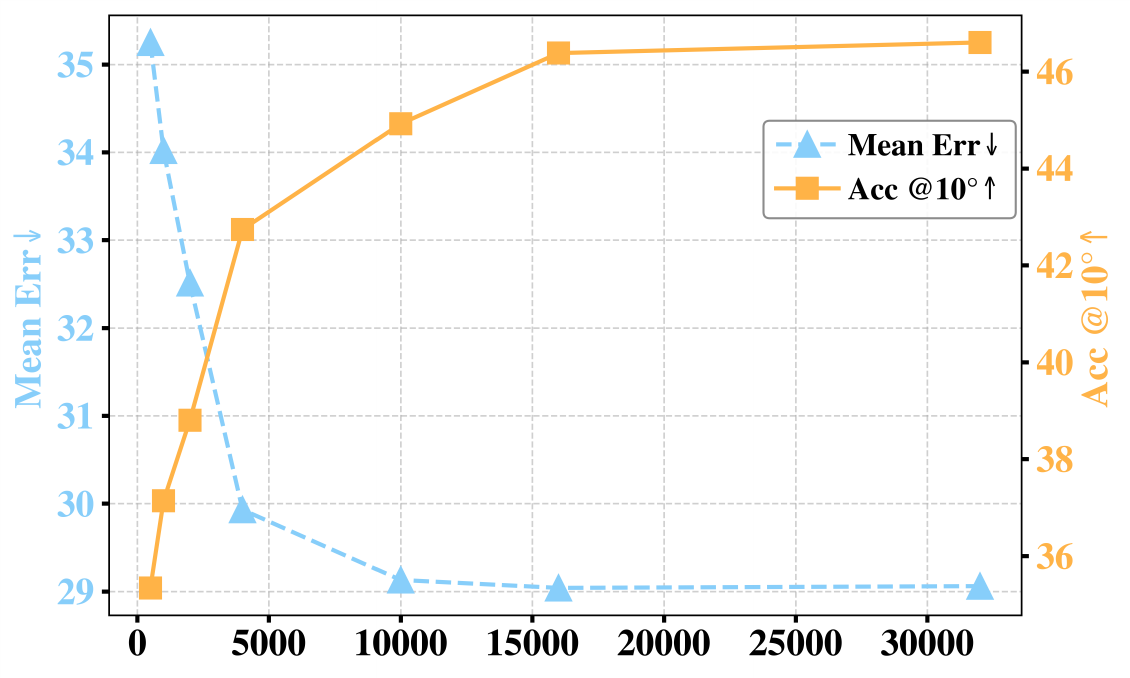}}
\vspace{-0.35cm}
\caption{\textbf{Ablation analysis on different number of uniformly initialized samples.} We tested sampling numbers of 500, 1000, 2000, 4000, 10000, 16000, 32000 on Mean Error and Acc @ $10^{\circ}$.}
\label{fig:fig_samples}
\vspace{-0.4cm}
\end{figure}

Moreover, we also perform extensive ablations on the sampling densities using uniform initialization. As shown in Fig. \ref{fig:fig_samples}, the performance boosts from 0 to 4000 samples, marginally improves from 4000 to 16000, and saturates after 16000 samples. In our experiments, we choose \emph{uniform initialization with 4000 samples} to balance the performance and the efficiency.

\begin{table}[!t]
\vspace{-0.35cm}
\renewcommand{\arraystretch}{1.2} 
  \caption{\textbf{Effects of different initialization strategies using the LineMOD dataset}.}
  \vspace{-0.1cm}
  \label{tab:ablation_initial}
  \centering
  \resizebox{\linewidth}{!}{
  \begin{tabular}{l||c||c|cccc}
    \hline
    \hline
   \multirow{2}{*}{Initial Strategy} & \multirow{2}{*}{Sampling Numbers} & Error↓ & \multicolumn{4}{c}{Acc @ $t^{\circ}$ (\%) ↑}\\
   \cline{3-7}
     &&Mean Err & 30$^{\circ}$ & 15$^{\circ}$ & 10$^{\circ}$ & 5$^{\circ}$ \\
    \hline
    Random Init. & 4000 & 31.73 &   70.90 & 52.66   & 40.88 & 23.08\\
    Uniform Init.  & 4000 &  29.93 & 72.06 & 54.90 & 42.74 & 24.32 \\
    \hline
    \hline
  \end{tabular}
}
\vspace{-0.2cm}
\end{table}

\begin{figure}[!t]
\centerline{\includegraphics[width=0.75\linewidth]{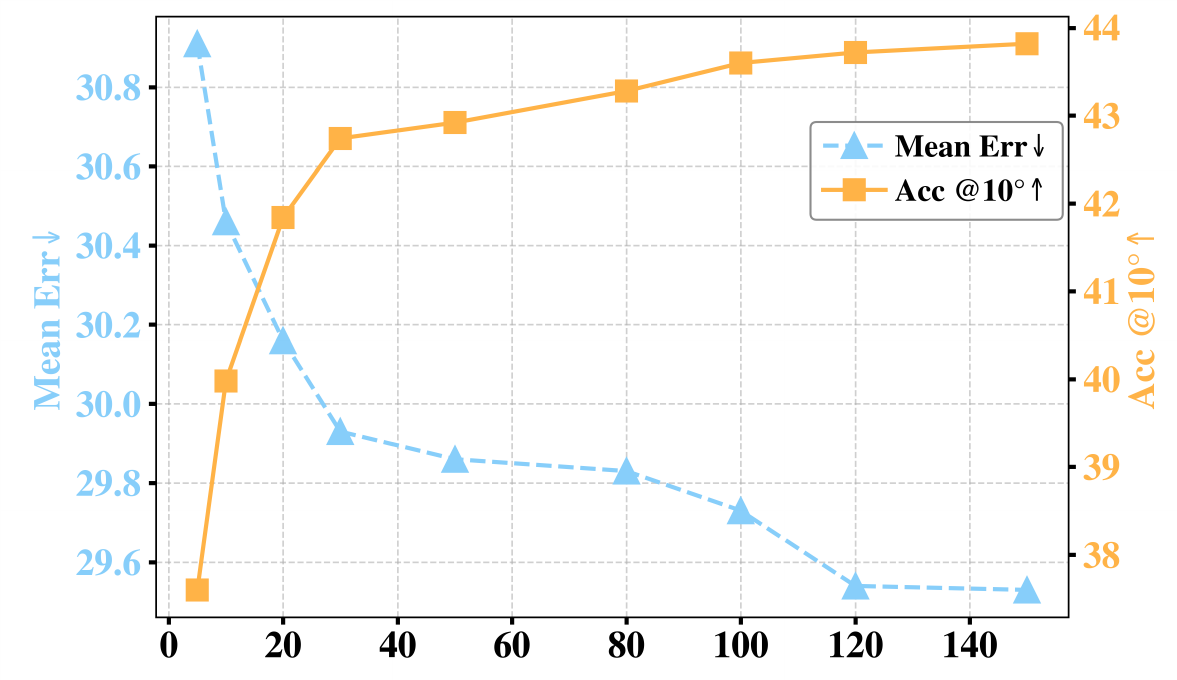}}
\vspace{-0.3cm}
\caption{\textbf{Performance trend w.r.t. number of refining iterations.} We tested iteration numbers of (5, 10, 20, 30, 50, 80, 100, 120, 150) on Mean Error and Acc @ $10^{\circ}$.}
\label{fig:fig_iteration}
\vspace{-0.3cm}
\end{figure}

\subsection{Effects of Different Refinement Iterations.} 
\label{sec:iter}
Figure \ref{fig:fig_iteration} illustrates the impact of the iteration numbers for our label/training-free refinement using the LineMOD dataset, which demonstrates that the performance boosts from 0 to 30 iterations, marginally improves from 30 to 120, and saturates after 120 iterations. We thus set the iteration number to 30 in our main experiments to achieve a balance between performance and efficiency.

\begin{table}[t]
\renewcommand{\arraystretch}{1.1} 
  \caption{\textbf{Inference time statistics} of our method on LineMOD.}
  \vspace{-0.2cm}
  \label{tab:time}
  \centering
  \resizebox{\linewidth}{!}{
  \begin{tabular}
  {cccc}
    \toprule
    Semantic Fea. Extraction & Pose Initialization & Refinement & Total \\
    \hline
    0.11s &  1.18s & 1.02s & 2.52s\\
  \bottomrule
  \end{tabular}
  }
\vspace{-0.3cm}
\end{table}

\begin{table}[!t]
\centering 
\renewcommand{\arraystretch}{1.1} 
\caption{Inference time comparison on LineMOD.} 
\vspace{-0.2cm}
\label{tab:time_comparison}
\resizebox{\linewidth}{!}{
\begin{tabular}{c|ccccc|c}
\hline
\hline
Method & ZSP & RelPose++ & LoFTR & 3DAHV & DMVNet & Ours
\\
\hline
Time & 1.72s & 0.69s & 0.30s & 0.04s & 0.04s & 2.52s \\
\hline
\hline
\end{tabular}
}
\vspace{-0.4cm}
\end{table}

\vspace{-0.1cm}
\subsection{Analysis on the Inference Time} 
\label{sec:time}
We collect the inference time per reference-query pair, averaged across the LineMOD datasets on a single 4090 GPU. We report the runtime for each stage of our method in Table \ref{tab:time}. Note that the initialization is efficient with much more candidate samples than the refinement, because those initializing candidate samples can be evaluated in parallel without backpropagation. Table \ref{tab:time} demonstrates the efficiency of our method with a per-pair runtime of 2.52 seconds in total.

We also present comparisons of our approach with the baseline methods, in terms of the inference time, in Table \ref{tab:time_comparison}. While our method is slower in inference than the state-of-the-art feedforward models, our render-and-compare paradigm is training-free. This eliminates the numerous training hours required by the state-of-the-art feedforward methods, and ensures inherent generalization to unseen objects (i.e., training-free brings desirable training-data independence).

\vspace{-0.2cm}
\subsection{Illustrations of the Failure Cases}
\label{sec:failure}

\begin{figure}[t]
\vspace{-0.3cm}
\centerline{\includegraphics[width=0.7\linewidth,height=0.32\linewidth]{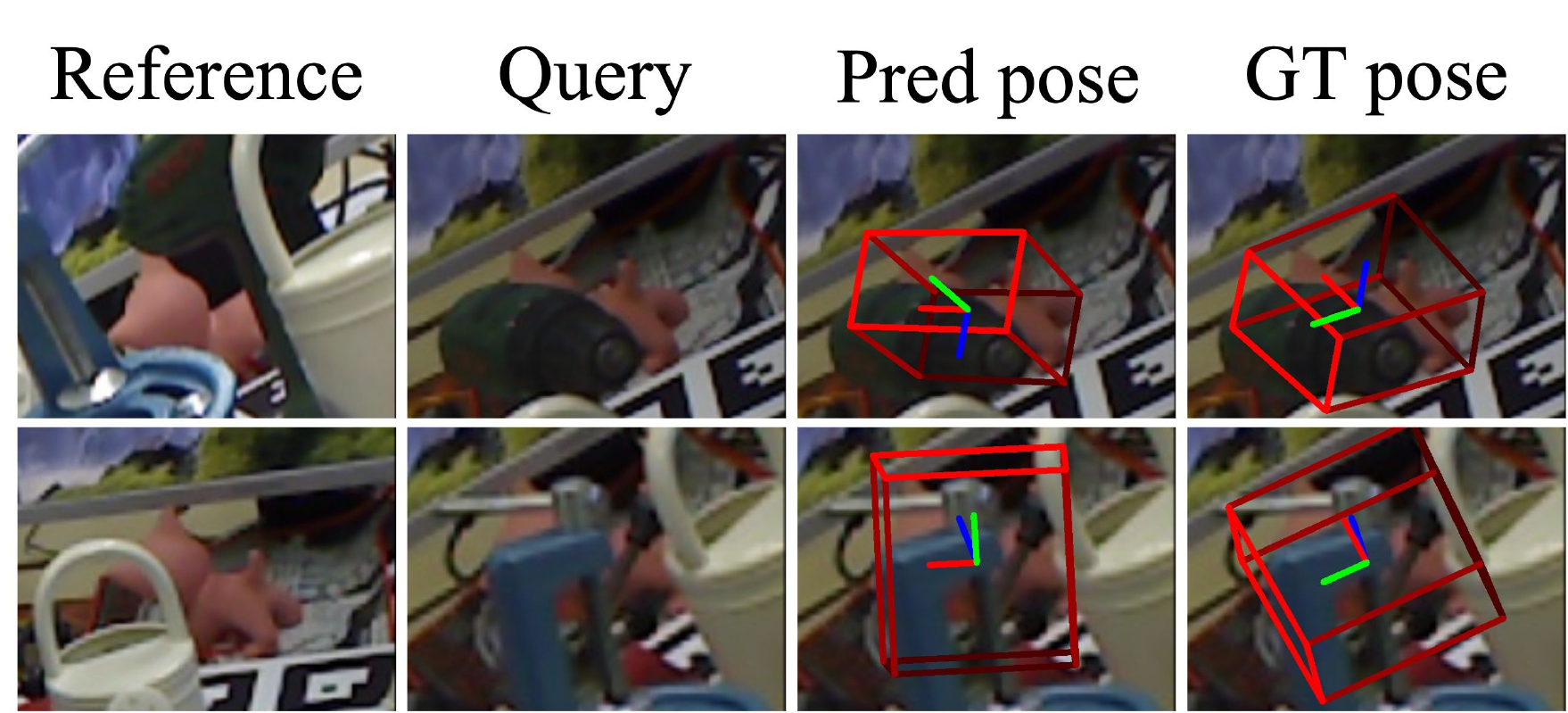}}
\caption{\textbf{Failure cases of our method on the LM-O dataset}, where there do not exist sufficient overlaps between the query and the reference due to severe occlusions.}
\label{fig:fig7}
\vspace{-0.3cm}
\end{figure}

We show our failure cases on the LM-O dataset in Fig. \ref{fig:fig7}, where there do not exist sufficient overlaps between the query and the reference. We note such an extremely degraded case as our limitation and discuss it in Sect. \ref{sec:conclusion} (Limitations and Future Works).

\section{Discussions and Conclusions} \label{sec:conclusion}

\noindent \textbf{Limitations and Future Works.} Our method has the following two limitations. Firstly, our method necessitates the depth information of the reference object as an input. To acquire the \textbf{metric} depth of the reference image, we evaluated a state-of-the-art monocular depth estimation algorithm \cite{yang2024depth}. The results and discussions in Sect. S3 of the supplementary shows that the metric depth estimation occasionally fails to generalize across different datasets, primarily stemming from the inherent metric scale ambiguities under varying camera parameters and diverse objects. This suggests that \textit{the current limitations are likely to be overcome once a generalizable \textbf{metric} depth estimator becomes available}. Despite this, we note that depth sensors are commonly used in our primary application domain, i.e., robotics. Our empirical results, presented in Table S1 of the supplementary materials, demonstrate that our method remains robust with imprecise depth obtained by a noisy depth sensor (simulated by adding noise to the ground-truth depth).

Secondly, our method is likely to fail in the severely degraded scenario where there do not exist adequate overlaps between the query and the reference (possibly caused by occlusions, e.g., Fig. \ref{fig:fig7}). Future research with simultaneous render-and-compare and object completion (with minimal inconsistent hallucination) is a promising direction to explore.

We also note an additional future direction about adaptively determining the loss weights of the RGB pair and the semantic pair in Eq. \eqref{eq:equation2} (preferably adapting in each refinement step), though we empirically showed that simply using equal weights (i.e., both set to 1) leads to promising results.

\noindent \textbf{Conclusions.} In this paper, we addressed the challenging generalizable relative pose estimation under a rigorous circumstance with only a single RGB-D reference and single RGB query pair as input, and the pose label is not a priori. We establish our label- and training-free method following the render-and-compare paradigm, by exploiting i) the 2.5D (i.e., RGB-D) rotatable reference mesh, ii) the semantic maps of both query and reference (extracted by a pretrained large vision model DINOv2), and iii) a differentiable renderer to produce and back-propagate losses to refine the relative pose. We carried out extensive experiments on the LineMOD, LM-O, and YCB-V datasets. The results demonstrate that our label/training-free approach surpasses the performance of state-of-the-art supervised methods, particularly excelling under the rigorous \texttt{Acc@5/10/15}$^\circ$ metrics.

{\footnotesize
\bibliographystyle{plain}
\bibliography{main}
}

\vspace{-12mm}
\begin{IEEEbiography}
[{\includegraphics[width=1.5in,height=1.25in,clip,keepaspectratio]{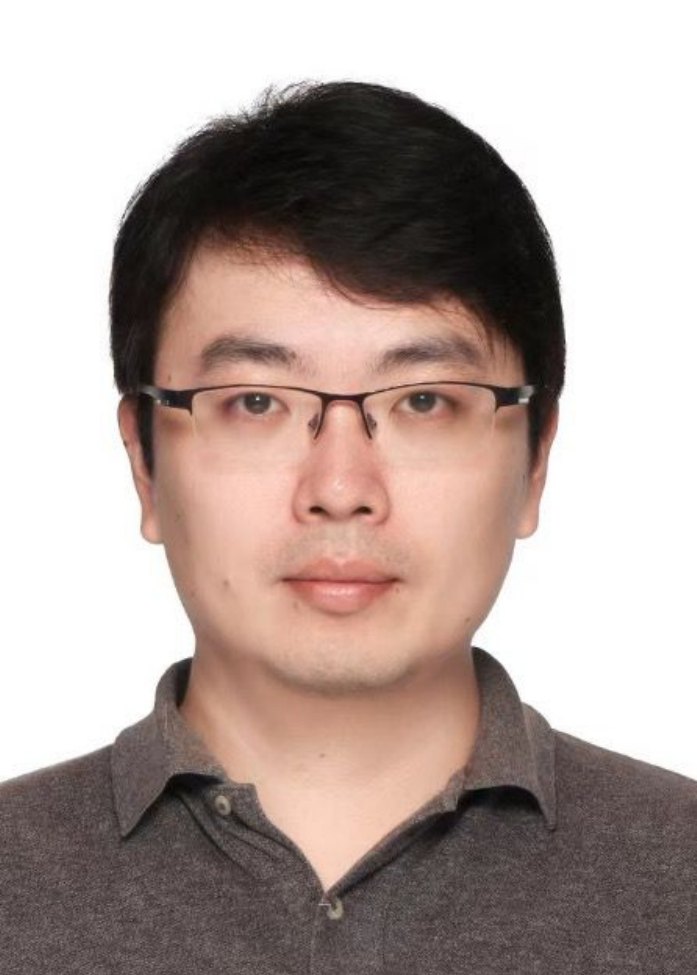}}]{Yuan Gao}
(Member, IEEE) received the B.S. degree and the M.S. degree from Huazhong University of Science and Technology, and the Ph.D. degree from City University of Hong Kong, in 2009, 2012, and 2016, respectively. He was a visiting graduate researcher with University of California, Los Angeles in 2015, and a senior research scientist with Tencent AI Lab from 2017 to 2020. Currently, he is an Associate Professor with School of Artificial Intelligence, Wuhan University. His research interests include 3D computer vision, multi-task/modal learning, and efficient deep learning.
\vspace{-10mm}
\end{IEEEbiography}

\begin{IEEEbiography}[{\includegraphics[width=1.5in,height=1.25in,clip,keepaspectratio]{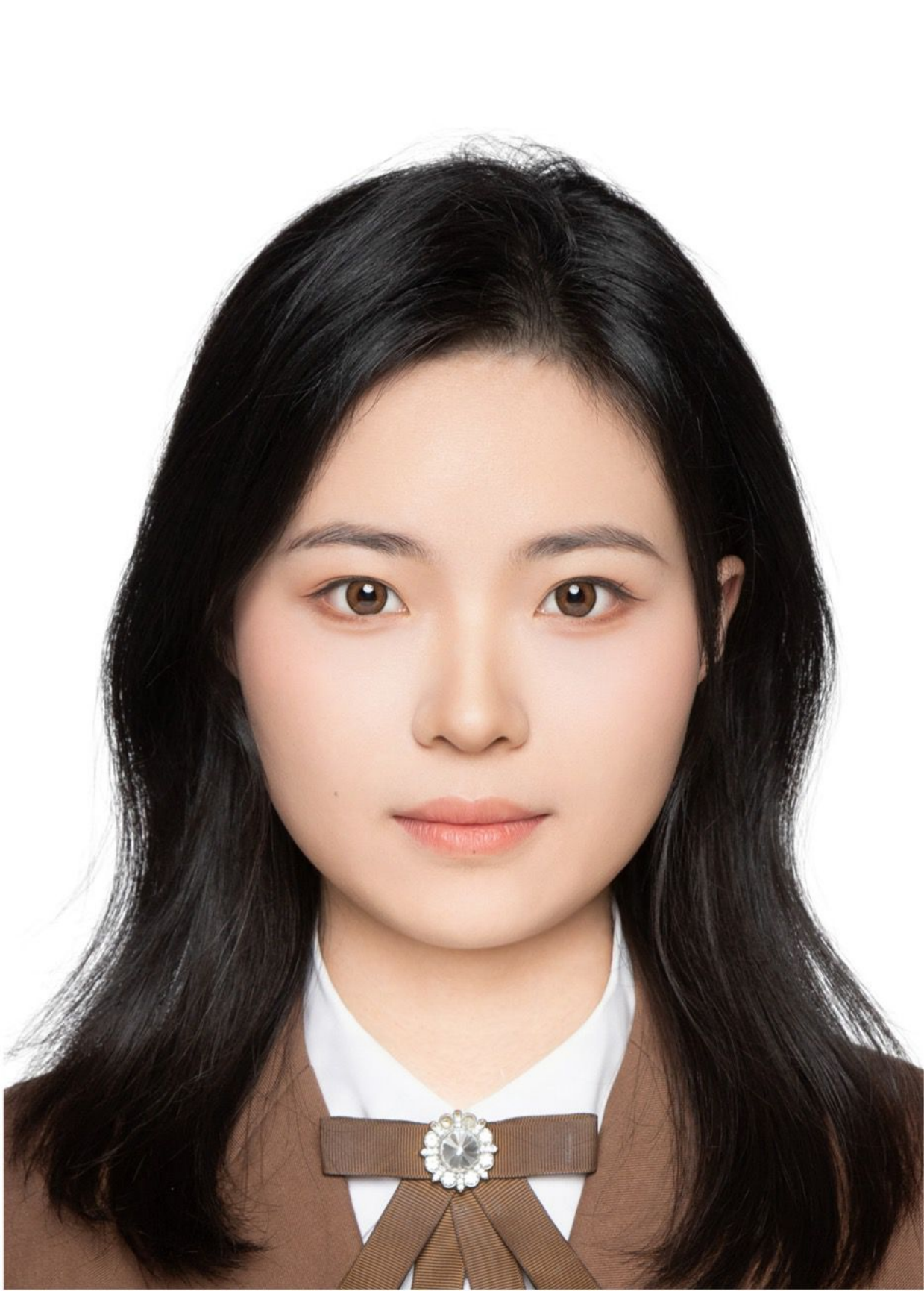}}]{Yajing Luo}
received the B.S. degree from the School of Computer Science, Wuhan University in 2022. She is currently pursuing the Ph.D. degree with the School of Computer Science, Wuhan University. Her research interests include object pose estimation and 3D scene understanding.
\vspace{-10mm}
\end{IEEEbiography}

\begin{IEEEbiography}[{\includegraphics[width=1.5in,height=1.25in,clip,keepaspectratio]{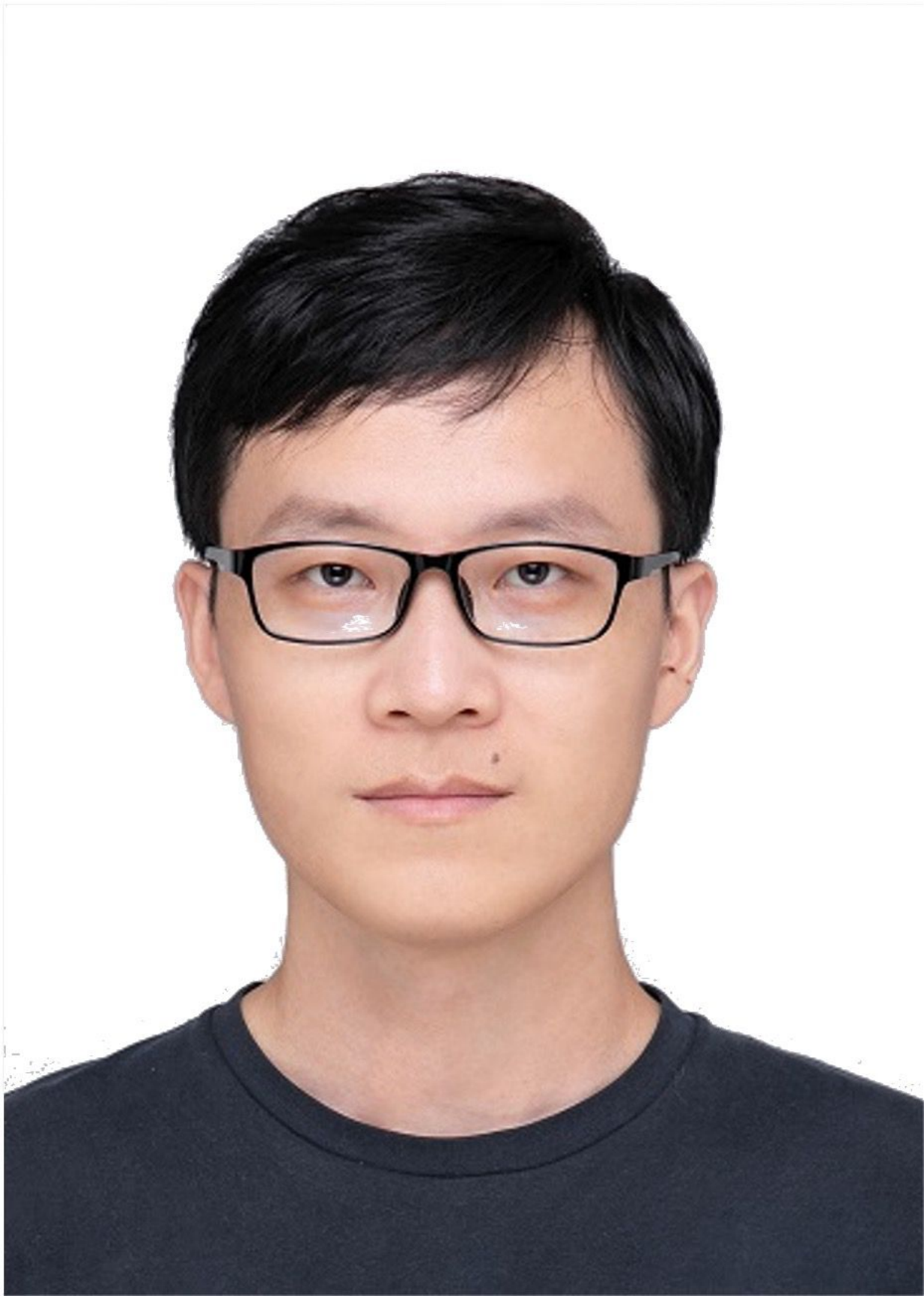}}]{Junhong Wang}
received the B.S. and M.S. degrees from Huazhong University of Science and Technology, Wuhan, China, in 2009 and 2012. He is now a graphics software engineer in Tencent Games since 2012. His research interests include 3D computer graphics and mobile rendering.
\vspace{-10mm}
\end{IEEEbiography}

\begin{IEEEbiography}[{\includegraphics[width=1.5in,height=1.25in,clip,keepaspectratio]{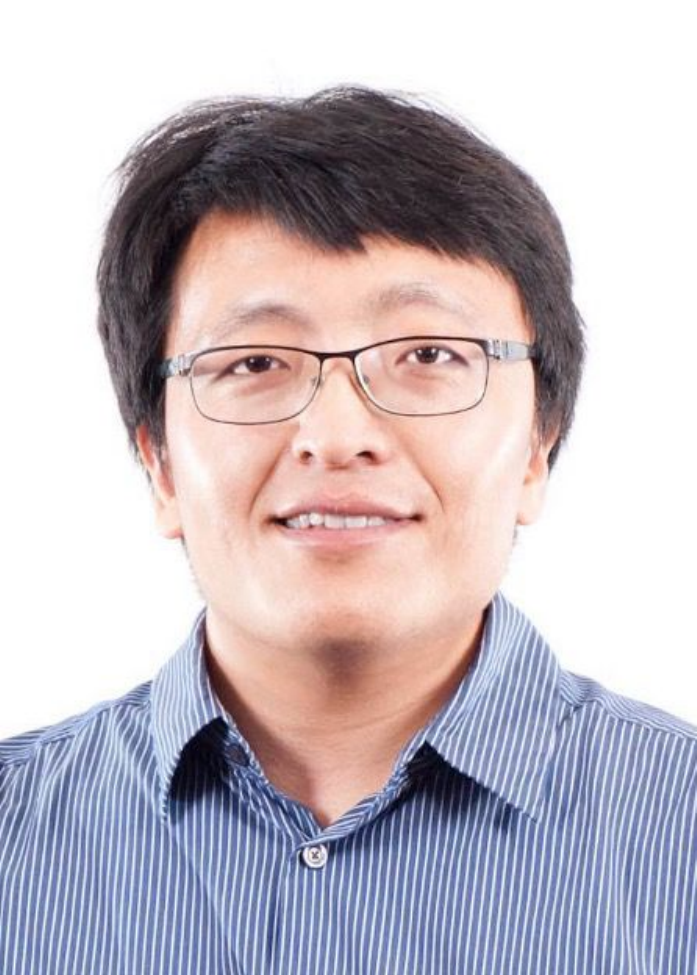}}]{Kui Jia}
(Member, IEEE) received the B.E. degree from Northwestern Polytechnic University, Xi’an, China, in 2001, the M.E. degree from the National University of Singapore, Singapore, in 2004, and the Ph.D. degree in computer science from the Queen Mary University of London, London, U.K., in 2007. He was with the Shenzhen Institute of Advanced Technology of the Chinese Academy of Sciences, Shenzhen, China, Chinese University of Hong Kong, Hong Kong, the Institute of Advanced Studies, University of Illinois at Urbana-Champaign, Champaign, IL, USA, the University of Macau, Macau, China, South China University of Technology, Guangzhou, China. He is currently a Professor with the School of Data Science, the Chinese University of Hong Kong, Shenzhen, China. His recent research focuses on theoretical deep learning and its applications in vision and robotic problems, including deep learning of 3D data and deep transfer learning. He serves on the Editorial Boards of \emph{IEEE Transactions on Image Processing}, and \emph{Transactions on Machine Learning Research}.
\vspace{-10mm}
\end{IEEEbiography}

\begin{IEEEbiography}[{\includegraphics[width=1.5in,height=1.25in,clip,keepaspectratio]{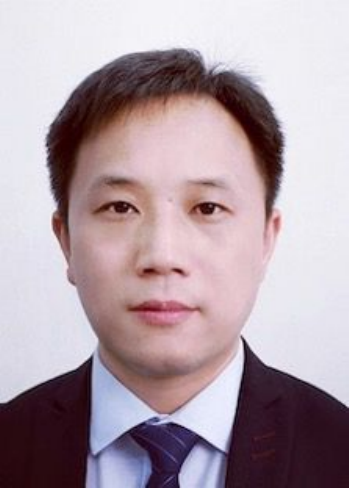}}]{Gui-Song Xia}
(Senior Member, IEEE) received the PhD degree in image processing and computer vision from CNRS LTCI, T\'el\'ecom ParisTech, Paris, France, in 2011. From 2011 to 2012, he was a postdoctoral researcher with the Centre de Recherche en Math\'ematiques de la Decision, CNRS, Paris Dauphine University, Paris, for one and a half years. He is currently working as a full professor in computer vision and photogrammetry with Wuhan University. He has also been working as a visiting scholar at DMA, \'Ecole Normale Sup\'erieure (ENS-Paris) for two months, in 2018. He is also a guest professor of the Future Lab AI4EO in Technical University of Munich (TUM). His current research interests include mathematical modeling of images and videos, structure from motion, perceptual grouping, and remote sensing image understanding. He serves on the Editorial Boards of several journals, including \emph{ISPRS Journal of Photogrammetry and Remote Sensing, Pattern Recognition, Signal Processing: Image Communications, EURASIP Journal on Image \& Video Processing, Journal of Remote Sensing}, and \emph{Frontiers in Computer Science: Computer Vision.}
\vspace{-10mm}
\end{IEEEbiography}

\clearpage 


\section*{Supplementary material}

# Supplementary Material for the Paper:

# Towards Human-level 3D Relative Pose Estimation: Generalizable, Training-Free, with Single Reference

Yuan Gao*, Yajing Luo*, Junhong Wang, Kui Jia, Gui-Song Xia

We address the following issues in the supplementary material files:

1) Angle Error Distribution on LM-O in Sect. S1.
2) Experiments with imprecise input depth in Sect. S2.
3) Our results on the LineMOD [3], LM-O [1], and YCB-V [6] datasets w.r.t. **per object** in Sect. S3.
4) Qualitative Results on the LineMOD [3], LM-O [1], and YCB-V [6] datasets for all the methods in Sect. S4.
5) We also attached a video depicting the overview and the label/training-free refinement procedure of our method (i.e., the video version of Fig. 2 in the main text), as well as the qualitative results.

## S1. ANGLE ERROR DISTRIBUTION ON THE LM-O DATASET.

Figure S1 presents the angle error distribution (ranging from 0 to 180 degrees) for all the methods on the LM-O dataset. The statistics reveal that at lower angle error thresholds (e.g., for $t \leq 10, 20$ in Acc@$t°$), our approach substantially outperforms both 3DAHV [8] and DVMNet [7].

## S2. EXPERIMENTS WITH IMPRECISE INPUT DEPTH

As discussed in Sec. I.A (Applicability) in the main text, our method has the potential to use imprecise depth. We validate this on the LineMOD [3] dataset. Concretely, we simulate the imprecise depth by adding Gaussian noise $\mathcal{N}(0, \sigma)$ to the ground-truth depth map $D_r$, where $\sigma$ is set to:

$$\sigma = \lambda * d, \qquad d = \max(D_r) - \min(D_r), \qquad (1)$$

where $d$ is the maximal depth difference of the input sample.

We validate different $\lambda$'s with 0.001, 0.003, and 0.005 on the LineMOD [3] dataset, the results are shown in Table S1, which demonstrate the potential of integrating our method with an off-the-shelf generalizable depth estimator.

Y. Gao is with the School of Electronic Information, Wuhan University, Wuhan, China. E-mail: ethan.y.gao@gmail.com

Y. Luo and G.-S. Xia are with the School of Computer Science, Wuhan University, Wuhan, China. E-mails: {yajingluo, guisong.xia}@whu.edu.cn

J. Wang is with MoreFun Studio, Tencent Games, Tencent, Shenzhen, China. E-mails: junhongwang@tencent.com

K. Jia is with the School of Data Science, The Chinese University of Hong Kong, Shenzhen, China. E-mails: kuijia@cuhk.edu.cn

Corresponding authors: Gui-Song Xia, Yuan Gao.

* indicates equal contribution.

TABLE S1
**EXPERIMENTS WITH IMPRECISE DEPTH ON LINEMOD.**

| Metrics | Mean Err↓ | Acc@30°↑ | Acc@15°↑ | Acc@10°↑ | Acc@5°↑ |
|---------|-----------|----------|----------|----------|---------|
| $\lambda = 0.005$ | 33.10 | 69.52 | 52.66 | 40.16 | 20.64 |
| $\lambda = 0.003$ | 30.92 | 71.44 | 54.96 | 42.90 | 23.10 |
| $\lambda = 0.001$ | 30.11 | 72.00 | 55.10 | 43.04 | 24.22 |
| Ours | **29.93** | **72.06** | **54.90** | **42.74** | **24.32** |

TABLE S2
**OUR RESULTS ON THE LINEMOD DATASET W.R.T. PER OBJECT.** THE OBJECTS WITH RED TEXT ARE THOSE USED FOR TESTING IN THE MAIN PAPER.

| Object | Mean Err↓ | Acc@30°↑ | Acc@15°↑ | Acc@10°↑ | Acc@5°↑ |
|--------|-----------|----------|----------|----------|---------|
| ape | 43.41 | 46.90 | 26.10 | 17.40 | 5.80 |
| benchvise | 17.79 | 87.30 | 75.60 | 64.60 | 42.10 |
| camera | 24.10 | 73.70 | 58.00 | 46.80 | 27.60 |
| can | 26.49 | 75.00 | 63.20 | 55.00 | 37.80 |
| cat | 33.90 | 68.00 | 52.20 | 40.20 | 22.00 |
| driller | 35.58 | 76.90 | 59.40 | 44.60 | 23.90 |
| duck | 38.30 | 54.40 | 29.30 | 17.50 | 6.00 |
| eggbox | 27.63 | 77.40 | 66.10 | 57.10 | 36.90 |
| glue | 46.35 | 55.30 | 38.20 | 28.00 | 13.70 |
| holepuncher | 26.25 | 76.50 | 64.30 | 52.30 | 26.80 |
| iron | 33.20 | 73.70 | 58.80 | 48.90 | 29.60 |
| lamp | 29.28 | 79.30 | 66.20 | 56.20 | 36.90 |
| phone | 25.82 | 80.80 | 64.60 | 50.20 | 27.30 |
| average | 31.39 | 71.17 | 55.54 | 44.52 | 25.88 |

## S3. PER OBJECT RESULTS ON THE LINEMOD, LM-O, AND YCB-V DATASETS

We present our results w.r.t. per object of the full LineMOD [3], LM-O [1], and YCB-V [6] datasets in Tables S2, S3, and S4, respectively. The experimental settings are the same as those in the main text, i.e., Tables II-IV.

Tables S2, S3, and S4 show that our method performs well on all the objects of the three datasets without training, further validating the strong zero-shot unseen-object generalize-ability of our label/training-free method.

## S4. QUALITATIVE RESULTS ON THE LINEMOD, YCB-V, AND LM-O DATASETS

Qualitative results on the LineMOD [3], LM-O [1], and YCB-V [6] datasets are illustrated in Figs. S2, S3, and S4, respectively. The ground truth and predicted poses are visualized by axes and 3D bounding boxes.

As depicted in Figs. S2, S3, and S4, our method outperforms the state-of-the-art methods [2, 4, 5, 7, 8] qualitatively in all the three datasets.

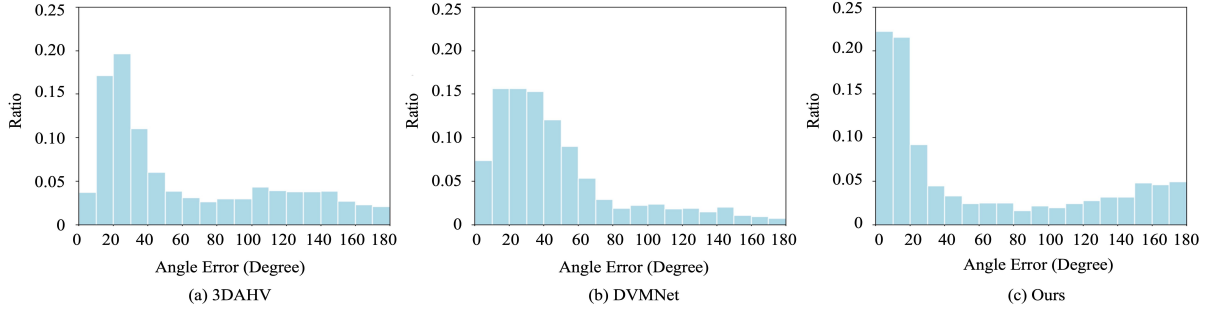

Fig. S1. **Angle Error Distribution on LM-O.**

TABLE S4

**OUR RESULTS ON THE YCB-V DATASET W.R.T. PER OBJECT.** THE OBJECTS WITH RED TEXT ARE THOSE USED FOR TESTING IN THE MAIN PAPER.

| Object | Mean Err↓ | Acc@30°↑ | Acc@15°↑ | Acc@10°↑ | Acc@5°↑ |
|---|---|---|---|---|---|
| 002_master_chef_can | 59.81 | 42.40 | 31.40 | 19.80 | 9.40 |
| 003_cracker_box | 83.20 | 40.40 | 32.70 | 26.80 | 8.50 |
| 004_sugar_box | 44.93 | 68.00 | 62.30 | 55.70 | 30.10 |
| 005_tomato_soup_can | 61.65 | 36.50 | 25.80 | 19.70 | 9.40 |
| 006_mustard_bottle | 20.76 | 86.40 | 83.30 | 77.80 | 59.70 |
| 007_tuna_fish_can | 111.75 | 16.50 | 10.20 | 8.10 | 4.70 |
| 008_pudding_box | 11.14 | 95.20 | 74.30 | 63.80 | 33.90 |
| 009_gelatin_box | 8.13 | 99.50 | 87.20 | 78.30 | 32.40 |
| 010_potted_meat_can | 99.01 | 26.80 | 22.30 | 13.20 | 4.60 |
| 011_banana | 46.94 | 56.40 | 46.00 | 33.60 | 12.30 |
| 019_pitcher_base | 33.43 | 58.60 | 41.40 | 28.50 | 9.50 |
| 021_bleach_cleanser | 51.91 | 55.80 | 41.00 | 29.70 | 12.00 |
| 024_bowl | 17.89 | 90.50 | 63.80 | 35.20 | 11.10 |
| 025_mug | 43.98 | 40.80 | 22.90 | 15.50 | 3.30 |
| 035_power_drill | 42.45 | 67.60 | 48.90 | 32.50 | 13.10 |
| 036_wood_block | 39.52 | 69.10 | 48.50 | 30.50 | 10.50 |
| 037_scissors | 27.55 | 83.50 | 60.00 | 41.10 | 15.40 |
| 040_large_marker | 52.96 | 25.40 | 10.00 | 5.00 | 1.80 |
| 051_large_clamp | 71.20 | 43.10 | 26.60 | 18.80 | 7.70 |
| 052_extra_large_clamp | 88.61 | 27.00 | 14.90 | 7.40 | 2.80 |
| 061_foam_brick | 27.89 | 81.90 | 73.00 | 50.10 | 13.30 |
| average | 49.74 | 57.69 | 44.12 | 32.91 | 14.55 |

TABLE S3

**OUR RESULTS ON THE LM-O DATASET W.R.T. PER OBJECT.** THE OBJECTS WITH RED TEXT ARE THOSE USED FOR TESTING IN THE MAIN PAPER.

| Object | Mean Err↓ | Acc@30°↑ | Acc@15°↑ | Acc@10°↑ | Acc@5°↑ |
|---|---|---|---|---|---|
| ape | 60.85 | 42.10 | 22.40 | 11.60 | 2.10 |
| can | 38.49 | 63.60 | 53.10 | 41.70 | 19.60 |
| cat | 55.84 | 53.70 | 33.00 | 22.30 | 8.60 |
| driller | 52.29 | 59.10 | 42.20 | 28.50 | 7.30 |
| duck | 57.13 | 50.70 | 29.70 | 18.20 | 4.60 |
| eggbox | 45.94 | 58.90 | 43.40 | 34.50 | 16.00 |
| glue | 61.52 | 46.50 | 30.70 | 19.20 | 9.20 |
| holepuncher | 42.51 | 62.20 | 47.20 | 33.00 | 10.40 |
| average | 51.82 | 54.59 | 37.72 | 26.13 | 9.73 |

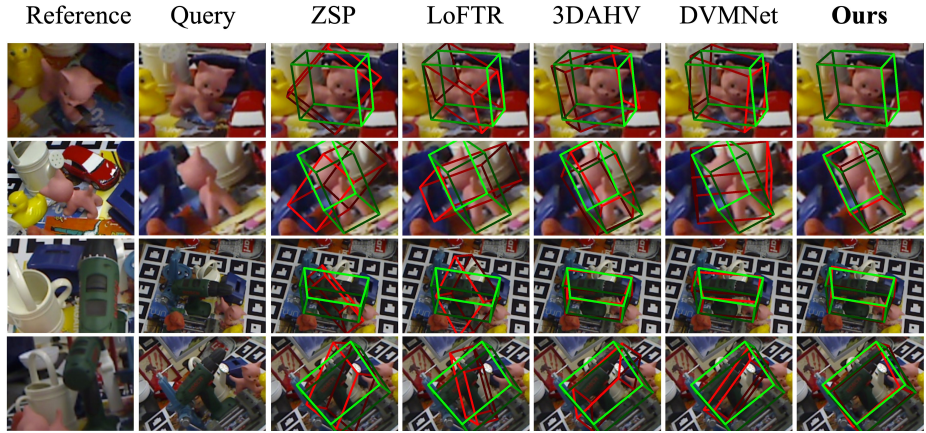

Fig. S2. **Qualitative results on LineMOD.** RelPose++ [4] did not release the LineMOD weights, the results of it in our main text were pasted from the 3DAHV paper [8], therefore the visualized results of RelPose++ are not included.

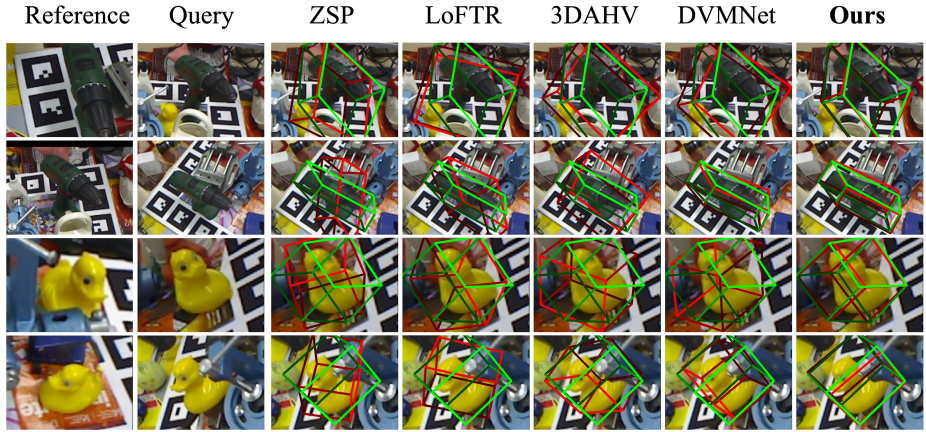

Fig. S3. **Qualitative results on LM-O.** LM-O is typically used solely to evaluate the models trained on LineMOD, since RelPose++ [4] have not released the LineMOD weights, the visualized results of RelPose++ are not included.

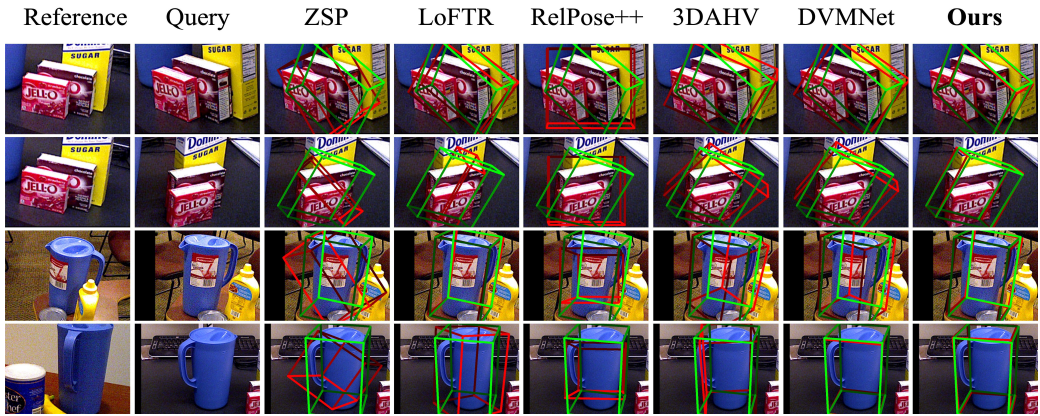

Fig. S4. **Qualitative results on YCB-V.** The predicted poses are visualized by <span style="color:red">red</span> 3D bounding boxes while the ground truth poses are depicted by <span style="color:green">green</span> 3D bounding boxes.



\end{document}